\pdfoutput=1
\documentclass[journal]{IEEEtran}
\usepackage[font=footnotesize,labelfont=bf]{caption}
\usepackage{epsfig,multirow}
\usepackage{amsthm,amsmath,amssymb,mathrsfs,bm}
\usepackage{amsfonts,dsfont,color,bbm}
\usepackage{algorithm,algorithmic}
\usepackage{subcaption}
\usepackage{cite,epstopdf,epsf,epsfig}
\usepackage{resizegather}
\usepackage{booktabs}
\usepackage{float}
\usepackage{mathtools}
\usepackage{enumitem}
\usepackage{comment}

\theoremstyle{break}

\newtheorem{remark}{Remark}

\theoremstyle{definition}

\usepackage{balance}

\begin{document}
\renewcommand{\thepage}{}

\title{Markovian RNN: An Adaptive Time Series Prediction Network with HMM-based Switching for Nonstationary Environments}

\author{Fatih Ilhan, Oguzhan Karaahmetoglu, Ismail Balaban and Suleyman S. Kozat, \textit{Senior Member, IEEE}
\thanks{This work in by part supported by Turkish Academy of Sciences Outstanding Researcher program.}
\thanks{F. Ilhan, O. Karaahmetoglu and S. S. Kozat are with the Department of Electrical and Electronics Engineering, Bilkent University, Ankara 06800, Turkey, e-mail: \{filhan, koguzhan, kozat\}@ee.bilkent.edu.tr}
\thanks{F. Ilhan, O. Karaahmetoglu, I. Balaban and S. S. Kozat are also with DataBoss A. S., Bilkent Cyberpark, Ankara 06800, Turkey, email: \{fatih.ilhan,oguzhan.karaahmetoglu,ismail.balaban,serdar.kozat\}@data-boss.com.tr}
}

\maketitle
\begin{abstract}
We investigate nonlinear regression for nonstationary sequential data. In most real-life applications such as business domains including finance, retail, energy and economy, time-series data exhibits nonstationarity due to the temporally varying dynamics of the underlying system. We introduce a novel recurrent neural network (RNN) architecture, which adaptively switches between internal regimes in a Markovian way to model the nonstationary nature of the given data. Our model, Markovian RNN employs a hidden Markov model (HMM) for regime transitions, where each regime controls hidden state transitions of the recurrent cell independently. We jointly optimize the whole network in an end-to-end fashion. We demonstrate the significant performance gains compared to vanilla RNN and conventional methods such as Markov Switching ARIMA through an extensive set of experiments with synthetic and real-life datasets. We also interpret the inferred parameters and regime belief values to analyze the underlying dynamics of the given sequences.
\end{abstract}

\section{Introduction}

\subsection{Preliminaries}

We study nonlinear time series prediction with recurrent neural networks in nonstationary environments. In particular, we receive a sequential data and predict the next samples of the given sequence based on the knowledge about history, which includes the previous values of target variables and side information (exogenous variables). Time series prediction task is extensively studied for various applications in the machine learning~\cite{Haykin,Tolga2018,lstm_}, ensemble learning~\cite{CBianchi2006,Vovk1990,Yuksel}, signal processing~\cite{Singer94,Engel2004,zhang}, and online learning theory~\cite{CBianchi2006,Denizcan2017} literatures. In most real-life scenarios such as finance and business applications, time-series data may not be an output of a single stochastic process since the environment can possess nonstationary behavior. In particular, the dynamics of the underlying system, which generates the given sequence can exhibit temporally varying statistics.  Moreover, the behavior of the system can even be chaotic or adversarial~\cite{Raginsky,miranian}. Therefore, successfully modeling the nonstationarity of the data carries importance while performing prediction.

Although linear models have been popular, partly since they have been integrated to most statistics and econometrics software packages, neural network-based methods are becoming widely preferred for the time series prediction task thanks to their ability to approximate highly nonlinear and complex functions~\cite{Shao14}. In particular, deep neural networks (DNNs) with multiple layers have been successful in resolving overfitting and generalization related issues. Although their success in some fields such as computer vision has been demonstrated in numerous studies~\cite{ObjDet}, this multi-layered structure is not suitable for sequential tasks since it cannot capture the temporal relations in time-series data properly~\cite{DeepRec13}. To this end, recurrent neural networks (RNNs) are used in sequential tasks thanks to their ability to exploit temporal behavior. RNNs contain a temporal memory called hidden state to store the past information, which helps them to model time-series more successfully in several different sequential tasks~\cite{Tolga2018,Liu16,Laptev2017,DeMulder2015,Lin96}. Hence, we consider nonlinear regression with RNN-based networks to perform time-series prediction.

To address the difficulties raised by nonstationarity, several methods mostly based on mixture of experts~\cite{zeevi,liehr,puma,ebrahimpour} are proposed due to their ability to represent nonstationary or piecewise sequential data. Mixture of experts models rely on the principle of divide and conquer, which aims to find the solution by partitioning the problem into smaller parts. These models usually consist of three main components: separate regressors called experts, a gate that separates the input space into regions, and a probabilistic model that combines the experts~\cite{Yuksel}. The fact that mixture of experts simplifies complex problems by allowing each expert to focus on specific parts of the problem with soft partitioning provides an important advantage while modeling nonstationary sequences~\cite{Yuksel}. However, these methods require training multiple experts separately, which disallows joint optimization and end-to-end training. In addition, these methods rely on enough diversity among experts such that each expert makes errors at different regions in the feature space. Otherwise, their performance compared to single expert models becomes negligibly better or even worse, if none of the experts can fit the data well enough~\cite{diversity}.

In certain studies, simple linear regressors such as autoregressive integrated moving average (ARIMA) models are considered as experts, where each individual expert specializes in a very small part of the problem~\cite{zeevi}. To perform gating operation between experts, several adaptive switching approaches such as Markovian switching and transition diagrams are widely preferred~\cite{ephraim,nystrup,kozat1,kozat2,liehr,changjin}. For instance, Markov switching models are particularly popular techniques for the time series prediction task, particularly in econometrics literature~\cite{hamilton,kns,tvtp,liehr,changjin,markovbook}. These approaches employ multiple linear regressors (or classifiers), where each regressor is responsible for characterizing the behavior of the time series in a different regime. The switching mechanism between these regimes is controlled by a first-order Markov chain. Markov switching model can capture more complex dynamic patterns and nonstationarity, especially when the assumption of the existence of different regimes with Markovian transitions hold. This model and its variants are applied in analyzing and forecasting business, economic and financial time series~\cite{nystrup,liehr}. For instance, these models are used to identify business cycles, which consist of several regimes such as expansion and recession states~\cite{kns}. However, none of these methods consider nonlinear regression with recurrent neural networks, which limits their capability to capture complex temporal patterns. Our model, Markovian RNN can be interpreted as a temporally adaptive mixture of experts model, where the regime-specific hidden state transition weights inside the RNN cell are employed as experts and HMM-based Markovian switching performs the gating operation between regimes. In this way, Markovian RNN can detect different regimes and focus on each of them separately through learning separate weights, which enables our model to adapt nonstationarity while making predictions.

Although there exists a significant amount of prior work on the time series prediction task in nonstationary environments, we, for the first time in the literature, utilize the benefits of recurrent neural networks and HMM-based switching for nonlinear regression of nonstationary sequential data. In this study, we introduce a novel time series prediction network, Markovian RNN, which combines the advantages of recurrent neural networks and Markov switching. Our model employs a recurrent neural network with multiple hidden state transition weights, where each weight corresponds to a different regime. We control the transitions between these regimes with a hidden Markov model, which models the regimes as part of a Markov process. In this way, our model can capture the complex sequential patterns thanks to RNNs, and handle nonstationary with Markovian switching. We also optimize the whole network jointly at single stage. Our model can also be extended using different RNN structures such as long short-term memory (LSTM)~\cite{lstm} and gated rectified unit (GRU)~\cite{gru} networks as remarked in Section \ref{sec:main_hmm}. Furthermore, we demonstrate the performance gains of Markovian RNN in the time series prediction task with respect to the conventional forecasting methods such as ARIMA with Markov switching~\cite{hamilton} and neural network-based methods such as vanilla RNN~\cite{Elman90}. We perform extensive experiments over both synthetic and real datasets. We also investigate the inferred regime beliefs and transitions, as well as analyzing forecasting error in terms of mean squared error.

\subsection{Prior Art and Comparisons}

A considerable amount of research has been conducted in machine learning, signal processing and econometrics literatures to perform time series prediction in nonstationary environments~\cite{liehr,Raginsky,Haykin}. Although there are widely embraced linear methods such as autoregression (AR), moving average (MA), autoregressive integrated moving average (ARIMA) and their variants in the conventional time series prediction framework, these methods fail to capture complex temporal patterns~\cite{Specht1991,Engel2004}, since they cannot fully capture nonlinearity. There exists a wide range of nonlinear approaches to perform regression in the machine learning and signal processing literatures~\cite{Haykin,Engel2004,Lin96}. However, these earlier methods suffer from practical disadvantages related to computation and memory. Furthermore, they can perform badly due to stability and overfitting issues~\cite{Singer94}. 

To overcome these limitations, neural network-based methods have been increasingly popular thanks to the developments in optimization and neural network literatures~\cite{Gou2007,Specht1991}. Most of the recent studies adopt recurrent neural networks and its variants for sequential tasks. Certain studies have successfully applied recurrent neural networks for language, speech and text processing tasks~\cite{Haykin,Narendra90}. This approach has also been used for anomaly detection in temporal sequences and time series classification~\cite{tolga19,yang}. In this study, we employ RNNs considering their power to capture complex nonlinear temporal patterns and generalization capability over unseen data.

There are several studies, which adopt mixture of experts based approaches for time series prediction. A mixture of ARMA experts is considered in \cite{zeevi} to obtain a universal approximator for prediction in stationary sequences. In the work, the authors interpret the mixture of experts as a form of neural network. Various studies have developed universal sequential decision algorithms by dividing sequences into independent segments and performing switching between linear regressors~\cite{kozat1,kozat2,zeevi}. However, since these works utilize linear regressors as experts, they perform poorly in challenging scenarios, particularly when the temporal patterns in the given sequence is more complex. To remedy these issues, mixture of heterogenous experts is proposed in~\cite{puma} for nonstationary sequences. Another study also employs nonlinear regressors as experts for stock price prediction task~\cite{ebrahimpour}. However, the nonlinear regressors employed in these studies have multi-layered perceptron architectures without any temporally recurrent structure. This layered structure poorly performs in capturing time dependencies of the data due to its lack of temporal memory. Therefore, DNNs provide limited performance in processing temporal data and modeling time series~\cite{DeepRec13}. Instead, we employ RNNs to handle the temporal patterns in time series data. In addition, we jointly optimize the whole network at single stage whereas mixture of experts models require separate training sessions for each expert.

Designing the gating model in mixture of experts approach is as crucial as choosing the expert models. For instance, authors employ randomized switching mechanism based on transition diagrams in \cite{kozat1,kozat2}. In another study, authors use a gating procedure based on a fuzzy inference system in \cite{ebrahimpour}. However, most studies, especially in the business domain and finance literature, prefer Markovian switching based approaches since they express financial cycles more accurately~\cite{kim_book}. Certain earlier variants of this approach such as Hamilton model~\cite{hamilton}, Kim-Nelson-Startz model~\cite{kim} and Filardo model~\cite{tvtp} have been commonly preferred and specifically designed for the tasks in business and finance applications~\cite{kim_book}. However, these statistical methods are not flexible in terms of modeling, since they employ linear models with certain assumptions such as sequences with varying mean and variance. In another study, a mixture of linear experts approach with hidden Markov gating model is proposed and trained with expectation-maximization (EM) algorithm~\cite{liehr}. Similar approaches have been applied for anomaly detection tasks as well. For instance, in \cite{cao}, authors develop an adaptive HMM with an anomaly state to detect price manipulations. Although Markovian switching-based methods are commonly used for sequential tasks in nonstationary environments, few of them consider nonlinear models, which are mostly simple multi-layer networks. In addition, they usually require multiple training sessions and cannot be optimized jointly. However, we introduce a jointly optimizable framework, which can utilize the benefits of nonlinear modeling capability of RNNs and adaptive Markovian switching with HMMs in an effective way.

\subsection{Contributions}

Our main contributions are as follows:

\begin{enumerate}
    \item For the first time in the literature, we introduce a novel time series prediction model, Markovian RNN, based on recurrent neural networks and regime switching controlled by HMM. This approach enables us to combine the modeling capabilities of RNNs and adaptivity obtained by HMM-based switching to handle nonstationarity.
    \item We use gradient descent based optimization to jointly learn the parameters of the proposed model. The proposed sequential learning algorithm for Markovian RNN enables us to train the whole network end-to-end at single stage with a clean pipeline.
    \item Our model can prevent oscillations caused by frequent regime transitions by detecting and ignoring outlier data. The sensitivity of the introduced model can readily be tuned to detect regime switches depending on the requirements of desired applications, or with cross-validation.
    \item Through an extensive set of experiments with synthetic and real-life datasets, we investigate the capability of Markovian RNN to handle nonstationary sequences with temporally varying statistics. We compare the prediction performance of the introduced model with respect to vanilla RNN and conventional methods with Markovian switching in terms of root mean squared error (RMSE) and mean absolute error (MAE). We also analyze the inferred regime beliefs and transitions to interpret our model.
\end{enumerate}

\subsection{Organization}

The organization of the paper is as follows. We define the time series prediction task and describe the framework that uses recurrent neural networks and hidden Markov models in Section \ref{sec:problem}. Then we provide the introduced model, switching mechanism, and sequential learning algorithm for Markovian RNN in Section \ref{sec:main}. In Section \ref{sec:sim}, we demonstrate the performance improvements of the introduced model over an extensive set of experiments with synthetic and real datasets and investigate the inferred regimes and transitions. Finally, we conclude the paper in Section \ref{sec:concl} with several remarks.

\section{Problem Description} 
\label{sec:problem}

In this paper, all vectors are column vectors and denoted by boldface lower case letters. Matrices are denoted by boldface upper case letters. $\boldsymbol{x}^T$ and $\mathbf{X}^T$ are the corresponding transposes of $\boldsymbol{x}$ and $\mathbf{X}$. $\|\boldsymbol{x}\|$ is the $\ell^2$-norm of $\boldsymbol{x}$. $\odot$ and $\oslash$ denotes the Hadamard product and division, i.e., element-wise multiplication and division, operations, respectively. $|\mathbf{X}|$ is the determinant of $\mathbf{X}$. For any vector $\boldsymbol{x}$, $x_i$ is the $i^{th}$ element of the vector. $x_{ij}$ is the element that belongs to $\mathbf{X}$ at the $i^{th}$ row and the $j^{th}$ column. $\text{sum}(\cdot)$ is the operation that sums the elements of a given vector or matrix. $\delta_{ij}$ is the Kronecker delta, which is equal to one if $i=j$ and zero otherwise. E-notation is used to express very large or small values such that $mEn = m \times 10^n$.

We study nonlinear prediction of nonstationary time series. We observe a vector sequence $\boldsymbol{x}_{1:T} \triangleq \{\boldsymbol{x}_t\}_{t=1}^T$, where $T$ is the length of the given sequence, and $\boldsymbol{x}_t \in \mathbb{R}^{n_x}$ is the input vector for the $t^{th}$ time step. Input vector can contain target variables (endogenous variables) as well as side information (exogenous variables). The target output signal corresponding to  $\boldsymbol{x}_{1:T}$ is given by $\boldsymbol{y}_{1:T} = \{\boldsymbol{y}_t\}_{t=1}^T$, where $\boldsymbol{y}_t \in \mathbb{R}^{n_y}$ is the desired output vector at the $t^{th}$ time step. Our goal is to estimate $\boldsymbol{y}_t$ using the inputs until the $t^{th}$ time step by
\noindent
\begin{align}
    \boldsymbol{\hat{y}}_t = f(\boldsymbol{x}_{1:t}; \boldsymbol{\theta}), \nonumber
\end{align}
\noindent
where $f$ is an adaptive nonlinear function parameterized with $\boldsymbol{\theta}$. After observing the target value $\boldsymbol{y}_t$, we suffer the loss $\ell(\boldsymbol{y}_t, \boldsymbol{\hat{y}}_t)$, and optimize the network with respect to this loss. We evaluate the performance of the network by the mean squared error obtained over the sequence with 
\noindent
\begin{equation}
    L_{\text{MSE}} = \frac{1}{T} \sum_{t=1}^T \ell_{\text{MSE}}(\boldsymbol{y}_t, \boldsymbol{\hat{y}}_t), 
\label{eq:loss_mse}
\end{equation}
\noindent
where
\noindent
\begin{align}
\ell_{\text{MSE}}(\boldsymbol{y}_t, \boldsymbol{\hat{y}}_t) = \boldsymbol{e}_t^T\boldsymbol{e}_t, \nonumber
\end{align}
and $\boldsymbol{e}_t \triangleq \boldsymbol{y}_t - \boldsymbol{\hat{y}}_t$ is the error vector at the $t^{th}$ time step. Other extensions are also possible such as mean absolute error (MAE) as remarked in Section \ref{sec:main_hmm}.

\begin{figure*}[t!]
    \centering
    \includegraphics[width=0.9\textwidth]{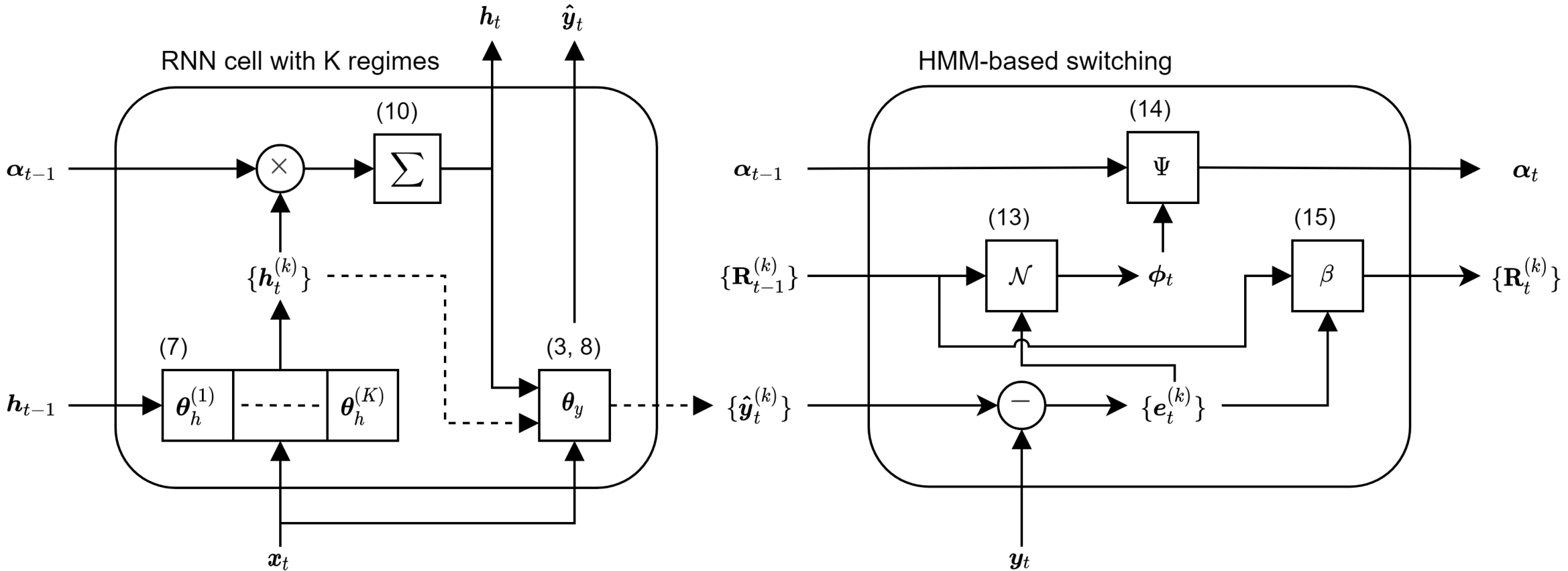}
    \caption{Detailed schematic of Markovian RNN cell. Here, $\boldsymbol{x}_t$, $\boldsymbol{y}_t$ and $\boldsymbol{\hat{y}}_t$ are the input, target and prediction vectors for the $t^{th}$ time step. $\boldsymbol{\alpha}_t$ and $\boldsymbol{h}_t$ are the belief state and hidden state vectors respectively. $\mathbf{R}^{(k)}_{t}$ is the error covariance matrix for the $k^{th}$ regime.}
    \label{fig:main}
\end{figure*}

\subsection{Recurrent Neural Networks}

We particularly study time series prediction with RNNs \cite{Williams89}. For this task, we use the following form:
\noindent
\begin{align}
\boldsymbol{h}_t &= f_h(\boldsymbol{h}_{t-1}, \boldsymbol{x}_t; \boldsymbol{\theta}_h) \nonumber \\ 
&= f_h(\mathbf{W}_{hh} \boldsymbol{h}_{t-1} + \mathbf{W}_{xh} \boldsymbol{x}_{t}) \label{eq:rnn_hh} \\
\boldsymbol{\hat{y}}_t &= f_y(\boldsymbol{h}_t; \boldsymbol{\theta}_y) \nonumber \\ 
&= f_y(\mathbf{W}_{hy} \boldsymbol{h}_t), \label{eq:rnn_hy}
\end{align}
\noindent
where $f_h(x) = \sigma_{\text{tanh}}(x)$ is the element-wise tanh function such that $\sigma_{\text{tanh}}(x) = \frac{e^x-e^{-x}}{e^x+e^{-x}}$, and $f_y(x) = x$. Here, $\boldsymbol{h}_t \in \mathbb{R}^{n_h}$ is the hidden state vector at time step $t$. $\mathbf{W}_{hh} \in \mathbb{R}^{n_h \times n_h}$, $\mathbf{W}_{xh} \in \mathbb{R}^{n_x \times n_h}$ and $\mathbf{W}_{hy} \in \mathbb{R}^{n_h \times n_y}$ are the weight matrices. We use $\boldsymbol{\theta}_h = \{\mathbf{W}_{hh}, \mathbf{W}_{xh}\}$ and $\boldsymbol{\theta}_y = \{\mathbf{W}_{hy}\}$ to denote the state transition and state-to-observation parameters respectively.


We note that the introduced framework can be applied for any stateful neural network structure. Hence, it can be extended to various RNN-based networks such as Gated Recurrent Units (GRU)~\cite{gru} and Long Short-term Memory Units (LSTM)~\cite{lstm}. We provide the equations for possible extensions in Remark \ref{remark:cell} in Section \ref{sec:main_regimes}. Here, we consider RNNs due to its simplicity and practicality in real-life applications. We also do not state bias terms explicitly since they can be augmented into input vector such that $\boldsymbol{x}_t \leftarrow [\boldsymbol{x}_t;\, 1]$.

\subsection{Hidden Markov Models}

We utilize HMMs to model the switching mechanism of RNNs, as will be described in Section \ref{sec:main}. HMM is a statistical model, which consists of a discrete-time discrete-state Markov chain with unobservable hidden states $k_t \in \{1, ..., K\}$, where $K$ is the number of states. The joint distribution has the following form:
\noindent
\begin{align}
    p(k_{1:T}, \boldsymbol{y}_{1:T}) = \prod\limits_{t=1}^{T}p(k_t|k_{t-1};\mathbf{\Psi})p(\boldsymbol{y}_t|k_t;\boldsymbol{\theta}),
\end{align}
\noindent
where $p(k_1|k_0) = \pi(k_1)$ is the inital state distribution. $p(k_t|k_{t-1};\mathbf{\Psi})$ is the transmission model defined by a transmission matrix $\mathbf{\Psi} \in \mathbb{R}^{K \times K}$ such that $\Psi_{ij} \triangleq p(k_t=j|k_t=i)$. The observation model (emission model) is sometimes modeled as a Gaussian such that $p(\boldsymbol{y}_t|k_t=k;\boldsymbol{\theta}) = \mathcal{N}(\boldsymbol{y}_t|\boldsymbol{\mu}_t, \mathbf{\Sigma}_t)$.

The state posterior $p(k_t|\boldsymbol{y}_{1:T})$ is also called the filtered belief state and can be estimated recursively by the forward algorithm~\cite{hmm_book} by
\noindent
\begin{align} \label{eq:hmm_forward}
    p(k_t|\boldsymbol{y}_t) &= \frac{p(\boldsymbol{y}_t|k_t)p(k_t|\boldsymbol{y}_{1:t-1})}{p(\boldsymbol{y}_t|\boldsymbol{y}_{1:t-1})} \nonumber \\
    &\propto p(\boldsymbol{y}_t|k_t)\sum\limits_{k_t=1}^K p(k_t|k_{t-1})p(k_{t-1}|\boldsymbol{y}_{t-1}). 
\end{align}
\noindent
Let $\alpha_{t, k} \triangleq p(k_t=k|\boldsymbol{y}_{1:T})$ denote the belief for the $k^{th}$ state, define $\boldsymbol{\alpha}_t = [..., \alpha_{t, k}, ...]^T$ as the $K$-dimensional belief state vector, and $\boldsymbol{\phi}_t = [..., p(\boldsymbol{y}_t|k_t=k), ...]^T$ as the $K$-dimensional likelihood vector respectively. Then \eqref{eq:hmm_forward} can be expressed as
\begin{align} \label{eq:hmm_forward_vec}
    \boldsymbol{\alpha}_t \propto \boldsymbol{\phi}_t \odot (\mathbf{\Psi}^T \boldsymbol{\alpha}_{t-1}).
\end{align}
\noindent
The filtered belief state vector can be obtained after normalizing the expression in \eqref{eq:hmm_forward_vec} through dividing by the sum of values. We note that we call HMM states as regimes from now on to prevent terminological ambiguity with the hidden states of RNN.

In the following section, we introduce the Markovian RNN architecture with HMM-based switching between regimes. We also provide the equations and the sequential learning algorithm of our framework.

\section{A Novel RNN Structure}
\label{sec:main}

In this section, we introduce our novel contributions for sequential learning with RNNs. We provide the structure of the Markovian RNN, by describing the modified network with multiple internal regimes in Section \ref{sec:main_regimes} and HMM-based switching mechanism in \ref{sec:main_hmm}. We present the sequential learning algorithm for the introduced framework in Section \ref{sec:main_opt}. The detailed schematic of the overall structure of our model is given in Fig. \ref{fig:main}.

\subsection{RNNs with Multiple Internal Regimes}
\label{sec:main_regimes}
Here, we describe the introduced Markovain RNN structure with multiple regimes, where each regime controls state transition independently. To this end, we modify the conventional form given in \eqref{eq:rnn_hh} and \eqref{eq:rnn_hy} as
\noindent
\begin{align}
\boldsymbol{h}^{(k)}_t &= f_h(\boldsymbol{h}_{t-1}, \boldsymbol{x}_t; \boldsymbol{\theta}^{(k)}_h) \nonumber \\ 
&= f_h(\mathbf{W}^{(k)}_{hh} \boldsymbol{h}_{t-1} + \mathbf{W}^{(k)}_{xh} \boldsymbol{x}_{t}), \label{eq:rnn_hh_k} \\
\boldsymbol{\hat{y}}^{(k)}_t &= f_y(\boldsymbol{h}^{(k)}_t; \boldsymbol{\theta}_y) \nonumber \\ 
&= f_y(\mathbf{W}_{hy} \boldsymbol{h}^{(k)}_t), \label{eq:rnn_hy_k}
\end{align}
\noindent
where $k \in \{1, ..., K\}$ is the regime index, and $K$ is the number of regimes. We also illustrate the modified RNN cell with multiple regimes in the left hand side of Fig. \ref{fig:main}. Here, the hidden state vector is independently propagated to the next time step at each node with different weights $\boldsymbol{\theta}^{(k)}_h$. We highlight that the state-to-observation model is same for all regimes. However, the resulting predictions $\boldsymbol{\hat{y}}^{(k)}_t$ are still different for each regime because of different hidden states $\boldsymbol{h}^{(k)}_t$ obtained for the $t^{th}$ time step.

We obtain the final estimate of the hidden state at time step $t$ by the weighted average of hidden states of each regime as
\noindent
\begin{align} \label{eq:hw_average}
    \boldsymbol{h}_t = \sum\limits_{k=1}^K w_{t, k}\boldsymbol{h}^{(k)}_t,
\end{align}
\noindent
where $w_{t, k}$ is the weight for the $k^{th}$ regime. Finally, we estimate the output using \eqref{eq:rnn_hy}. Here, the weights  $w_{t, k}$ are determined by the switching mechanism described in Section \ref{sec:main_hmm}. The number of states, $K$, is considered as a hyperparameter and can be selected using cross-validation.

\begin{remark}
\label{remark:cell}
Our model can also be extended with different RNN structures such as long short-term memory (LSTM)~\cite{lstm} and gated rectified unit (GRU)~\cite{gru}. For instance, for LSTM, all gating operations and state updates can be performed for each internal regime with the following equations:
\begin{align}
    &\boldsymbol{c}_t^{(k)} = \boldsymbol{c}_{t-1}^{(k)} \odot \boldsymbol{f}_t^{(k)} + \Tilde{\boldsymbol{c}}_t^{(k)} \odot \boldsymbol{i}_t^{(k)}, \nonumber \\
    &\boldsymbol{h}_t^{(k)} = f_h(\boldsymbol{c}_t^{(k)}) \odot \boldsymbol{o}_t^{(k)}, \nonumber \\
    &\boldsymbol{\hat{y}}^{(k)}_t = f_y(\mathbf{W}_{hy} \boldsymbol{h}^{(k)}_t) \nonumber 
\end{align}
where $\boldsymbol{f}_t^{(k)}$, $\boldsymbol{i}_t^{(k)}$, $\boldsymbol{o}_t^{(k)}$ are the forget, input and output gates, and $\Tilde{\boldsymbol{c}}_t^{(k)}$ is the candidate cell state at time step $t$ for the $k^{th}$ regime such that
\begin{align}
    &\boldsymbol{f}_t^{(k)} = \sigma(\mathbf{W}_{xf} \boldsymbol{x}_t + \mathbf{W}_{fh} \boldsymbol{h}_{t-1}^{(k)}), \nonumber \\ 
    &\boldsymbol{i}_t^{(k)} = \sigma(\mathbf{W}_{xi} \boldsymbol{x}_t + \mathbf{W}_{ih} \boldsymbol{h}_{t-1}^{(k)}), \nonumber \\
    &\Tilde{\boldsymbol{c}}_t^{(k)} = f_h(\mathbf{W}_{xg} \boldsymbol{x}_t + \mathbf{W}_{gh} \boldsymbol{h}_{t-1}^{(k)}), \nonumber \\
    &\boldsymbol{o}_t^{(k)} = \sigma(\mathbf{W}_{xo} \boldsymbol{x}_t + \mathbf{W}_{oh} \boldsymbol{h}_{t-1}^{(k)}),\nonumber
\end{align}
where $\sigma$ and $f_h$ are nonlinear element-wise activation functions. For the final estimates of the hidden state, we can apply \eqref{eq:hw_average}. For the cell state, the same form is applicable as well:
\begin{align}
    \boldsymbol{c}_t = \sum\limits_{k=1}^K w_{t, k}\boldsymbol{c}^{(k)}_t. \nonumber
\end{align}
The final output estimate $\hat{\boldsymbol{y}}_t$ can be calculated with \eqref{eq:rnn_hy}.
\end{remark}

\subsection{HMM-based Switching Mechanism}
\label{sec:main_hmm}
We employ an HMM to control the switching mechanism between internal regimes. In particular, we perform soft switching, where the weight given in \eqref{eq:hw_average} are represented using the belief values of the HMM as follows:
\noindent
\begin{align} \label{eq:h_average}
    \boldsymbol{h}_t = \sum\limits_{k=1}^K \alpha_{t-1, k}\boldsymbol{h}^{(k)}_t,
\end{align}
\noindent
where $\alpha_{t-1, k} \triangleq w_{t, k}$ denote the belief for the $k^{th}$ regime. To perform belief update as given in \eqref{eq:hmm_forward_vec}, we need to calculate the likelihood values of $\phi_t$ for the $t^{th}$ time step after observing $\boldsymbol{y}_t$. To this end, for mean squared error loss, we consider the error model with Gaussian distribution such that
\noindent
\begin{align} \label{eq:error_model}
    &\boldsymbol{y}_t = \boldsymbol{\hat{y}}^{(k)}_t + \boldsymbol{e}^{(k)}_t, \\
    &\boldsymbol{e}^{(k)}_t \sim \mathcal{N}(0, \mathbf{R}^{(k)}_{t-1}),
\end{align}
\noindent
where $\boldsymbol{e}^{(k)}_t$ is the error vector and $\mathbf{R}^{(k)}_{t-1}$ is the error covariance matrix for the $k^{th}$ regime, which stores the errors of the corresponding regime until the $t^{th}$ time step, excluding the last step. Then we compute the likelihood by
\noindent
\begin{align}
    p(\boldsymbol{y}_t|k_t=k) = \frac{1}{\sqrt{(2\pi)^{n_y} |\mathbf{R}_{t-1}^{(k)}|}} \exp{\left(-\frac{1}{2}\boldsymbol{e}^{{(k)}^T}_t \mathbf{R}_{t-1}^{{(k)}^{-1}} \boldsymbol{e}^{(k)}_t\right)}. \label{eq:likelihood}
\end{align}
\noindent
Once we obtain the likelihoods, we update the regime belief vector using \eqref{eq:hmm_forward_vec} as
\noindent
\begin{align} 
\begin{aligned} \label{eq:belief_update}
&\boldsymbol{\Tilde{\alpha}}_t = \boldsymbol{\phi}_t \odot (\mathbf{\Psi}^T \boldsymbol{\alpha}_{t-1}) \\
&\boldsymbol{\alpha}_t = \boldsymbol{\Tilde{\alpha}}_t / {\text{sum}(\boldsymbol{\Tilde{\alpha}}_t)},
\end{aligned}
\end{align}
\noindent
where we calculate $\boldsymbol{\phi}_t = [..., p(\boldsymbol{y}_t|k_t=k), ...]^T$ with \eqref{eq:likelihood}. We finally update the error covariance matrix using exponential smoothing by
\begin{align} \label{eq:R_update}
    \mathbf{R}^{(k)}_t = (1-\beta)\mathbf{R}^{(k)}_{t-1} + \beta\boldsymbol{e}^{(k)}_t\boldsymbol{e}_t^{{(k)}^T},
\end{align}
where $\beta \in [0, 1)$ controls the smoothing effect, which can be selected using cross validation. For instance, $\beta=0.95$ would result in high sensitivity to errors, which can cause outlier data to bring frequent oscillations between regimes, whereas very small values for $\beta$ might prevent the system to capture fast switches. The second part of the schematic in Fig. \ref{fig:main} illustrates the operations of HMM-based switching module.

\begin{remark}
\label{remark:loss}
Our frameworks can also be used with different loss functions such as mean absolute error (MAE) loss. In this case, we can model the distribution of the error vector $\boldsymbol{e}_t$ with the multivariate Laplacian distribution such that $\boldsymbol{e}_t \sim \mathcal{L}(0, \mathbf{\Sigma}_t)$, where $\mathbf{\Sigma}_t$ is the error covariance matrix at the $t^{th}$ time step. The likelihood computation of regimes given in \eqref{eq:likelihood} can be modified for the multivariate Laplacian distribution as
\begin{align}
    p(\boldsymbol{y}_t|k_t=k) &= \frac{2}{\sqrt{(2\pi)^{n_y} |\mathbf{R}_{t-1}^{(k)}|}} K_v(\sqrt{2}\rho) \left(-\frac{\rho}{2}\right)^{n_y/2}, \nonumber
\end{align}
where $\rho=\boldsymbol{e}^{{(k)}^T}_t \mathbf{R}_{t-1}^{{(k)}^{-1}} \boldsymbol{e}^{(k)}_t$, $v=1-n_y/2$ and $K_v$ is the modified Bessel function of the second kind~\cite{laplace}. For one-dimensional case ($n_y=1$), considering scalars instead of vectors, the likelihood equation reduces to
\begin{align}
    p(y_t|k_t=k) = \frac{1}{2r_{t-1}^{(k)}} \exp{\left(-\frac{|e_t^{(k)}|}{r_{t-1}^{(k)}}\right)}, \nonumber
\end{align}
where $e_t^k$ and $r_t^k$ are the error value and error variance at the $t^{th}$ time step for the $k^{th}$ regime.
\end{remark}
    
\begin{remark}
\label{remark:beta}
HMM-based switching inherently prevents instability due to the frequent oscillations between regimes or possible saturations at well-fitted regimes. One might argue that certain regimes that have been explored heavily during the early stages of the training would dominate other regimes and cause the system to degenerate into quite a few number of regimes. However, since the error covariance matrix penalizes errors for well-fit regimes more harshly than the regimes that are not explored yet, the model will tend to switch to other regimes if the predictions made by the dominating regimes start to produce high errors. Here, the choice of the smoothing parameter $\beta$ can be interpreted as an adjuster of the tolerance for the errors made in different regimes. As $\beta$ increases, the switching mechanism will have greater sensitivity to errors, which can cause instability and high deviations in the regime belief vector. Likewise, as $beta$ approaches towards zero, the system will not be able to capture switches due to the smoothing effect. This can eventually lead to saturations at well-fitted regimes. Thus, the choice of $\beta$ directly affects the behavior of our model and we can readily tune it depending on the needs of the specific application or select it with cross-validation. We further discuss and illustrate the effect of this parameter in Section \ref{sec:sim_analysis}.
\end{remark}

\subsection{Sequential Learning Algorithm for Markovian RNN}

\label{sec:main_opt}

In this section, we describe the learning algorithm of the introduced framework. During training, at each time step $t$, our model predicts the hidden state $\boldsymbol{h}_t$ and output $\boldsymbol{\hat{y}}_t$. We receive the loss given in \eqref{eq:loss_mse} after observing the target output $\boldsymbol{y}_t$. We denote the set of weights of our model as $\boldsymbol{\theta} = \{\{\boldsymbol{\theta}_h^{(k)}\}_{k=1}^K, \boldsymbol{\theta}_y, \mathbf{\Psi}\}$. We use the sequential gradient descent algorithm \cite{ZinkOGD} to jointly optimize the weights during the training.

In Algorithm \ref{alg:main}, we present the sequential learning algorithm for the introduced Markovian RNN. First, we initialize the model weights $\boldsymbol{\theta}$, hidden state $\boldsymbol{h}_1$, regime belief vector $\boldsymbol{\alpha}_1$ and error covariance matrices $\{\mathbf{R}_1^{(k)}\}_{k=1}^K$. For a given sequence with temporal length $T$, after receiving input $\boldsymbol{x}_t$ at each time step $t$, we compute hidden states for each internal regime using \eqref{eq:rnn_hh_k}. Then, we predict the output with these hidden states for each regime using \eqref{eq:rnn_hy_k}. After forward-pass of each internal regime, we generate $\boldsymbol{h}_t$ and output prediction $\boldsymbol{\hat{y}}_t$ using \eqref{eq:h_average} and \eqref{eq:rnn_hy}. After receiving the target output $\boldsymbol{y}_t$, we compute the loss using \eqref{eq:loss_mse} and update the model weights through the backpropagation of the derivatives. We use the Truncated Backpropagation Through Time (TBPTT) algorithm \cite{Will95}, and provide the derivatives of model weights in Appendix \ref{appendix}. To satisfy the requirement that each row of $\mathbf{\Psi}$ should sum to one, we scale the values row-wise using softmax function such that $\Psi_{ij} \leftarrow \frac{\exp{(\Psi_{ij})}}{\sum_{j'=1}^K\exp{(\Psi_{ij'})}}$. Finally, we update the regime belief vector and error covariance matrices by \eqref{eq:likelihood}-\eqref{eq:R_update}.

\begin{algorithm}[htbp!]
\algsetup{linenosize=\normalsize}
	\caption{Sequential Learning Algorithm for Markovian RNN}\label{alg:main}
	\begin{algorithmic}[1]
	    \STATE \textbf{Input:} Input and target time series: $\{\boldsymbol{x}_t\}_{t=1}^T$ and $\{\boldsymbol{y}_t\}_{t=1}^T$.
	    \STATE \textbf{Parameters:} Error covariance update term $\beta \in [0, 1)$, learning rate $\eta \in \mathbb{R}^+$, number of epochs $n \in \mathbb{N}^+$, early stop tolerance $n_{tolerance} \in \mathbb{N}$, training/validation set durations $T_{train}$ and $T_{val}$.
		\STATE \textbf{Initialize:} $\boldsymbol{\theta}$ (weights). \label{alg1:init}
	    \STATE \textbf{Initialize:} $\boldsymbol{\theta}_{best} = \boldsymbol{\theta}$ (best weights)
	    \STATE \textbf{Initialize:} $v = \infty$ (lowest validation loss)
	    \STATE \textbf{Initialize:} $j = 0$ (counter for early stop)
		\FOR{\textit{epoch} $e=1$ \TO $n$}
    		\STATE \textbf{Training Phase:}
    		\FOR{\textit{time step} $t=1$ \TO $T_{train}$}
	    		\STATE \textbf{Initialize:} $\boldsymbol{h}_1$, $\boldsymbol{\alpha}_1$ and $\{\mathbf{R}_t^{(k)}\}_{k=1}^K$.
        		\STATE \textbf{RNN Cell Forward Pass:}
        		\STATE Receive $\boldsymbol{x}_t$
        		\FOR{\textit{regime} $k=1$ \TO $K$}
        		\STATE $\boldsymbol{h}^{(k)}_t = f_h(\mathbf{W}^{(k)}_{hh} \boldsymbol{h}_{t-1} + \mathbf{W}^{(k)}_{xh} \boldsymbol{x}_{t})$
        		\STATE $\boldsymbol{\hat{y}}^{(k)}_t = f_y(\mathbf{W}_{hy} \boldsymbol{h}^{(k)}_t)$
        		\ENDFOR
        		\STATE $\boldsymbol{h}_t = \sum\limits_{k=1}^K \alpha_{t-1, k}\boldsymbol{h}^{(k)}_t$
        		\STATE $\boldsymbol{\hat{y}}_t = f_y(\mathbf{W}_{hy} \boldsymbol{h}_t)$
        		\STATE  \textbf{Calculate Loss:}
        		\STATE Receive $\boldsymbol{y}_t$
        		\STATE $\boldsymbol{e}_t = \boldsymbol{y}_t - \boldsymbol{\hat{y}}_t$
        		\STATE $\ell_{\text{MSE}}(\boldsymbol{y}_t, \boldsymbol{\hat{y}}_t) = \boldsymbol{e}_t^T\boldsymbol{e}_t$
        		\STATE \textbf{Backward Pass:}
        		\STATE Update model weights via bakcpropagation using $\frac{\partial \ell}{\partial \boldsymbol{\theta}}$
        		\STATE $\Psi_{ij} \leftarrow \frac{\exp{(\Psi_{ij})}}{\sum_{j'=1}^K\exp{(\Psi_{ij'})}}$
        		\STATE \textbf{HMM Based Switching:}
        		\STATE $\boldsymbol{\phi}_t = [..., p(\boldsymbol{y}_t|k_t=k), ...]^T$ from \eqref{eq:likelihood}
        		\STATE $\boldsymbol{\Tilde{\alpha}}_t = \boldsymbol{\phi}_t \odot (\mathbf{\Psi}^T \boldsymbol{\alpha}_{t-1})$
        		\STATE $\boldsymbol{\alpha}_t = \boldsymbol{\Tilde{\alpha}}_t / {\text{sum}(\boldsymbol{\Tilde{\alpha}}_t)}$
        		\STATE $\boldsymbol{e}_t^{(k)} = \boldsymbol{y}_t - \boldsymbol{\hat{y}}^{(k)}_t$
        		\STATE $\mathbf{R}^{(k)}_t = (1-\beta)\mathbf{R}^{(k)}_{t-1} + \beta\boldsymbol{e}^{(k)}\boldsymbol{e}^{{(k)}^T}$
    		\ENDFOR
			\STATE \textbf{Validation Phase:}
			\STATE $L_{val} = 0$ (validation loss)
		    \FOR{\textit{time step} $t=T_{train}$ \TO $T_{train}+T_{val}$}
    	        \STATE Make predictions $\boldsymbol{\hat{y}}_t$ using \eqref{eq:rnn_hy}-\eqref{eq:R_update}.
    	        \STATE $L_{val} = L_{val} + \ell_{\text{MSE}}(\boldsymbol{y}_t, \boldsymbol{\hat{y}}_t)$ 
    	    \ENDFOR
    	    \STATE $\bar{L}_{val} = \frac{L_{val}}{T_{val}}$
    	    \IF{$\bar{L}_{val} < v$}
    	        \STATE $v = \bar{L}_{val}$
    	        \STATE $\boldsymbol{\theta}_{best} = \boldsymbol{\theta}$
    	        \STATE $j = 0$
    	    \ELSE
    	        \STATE $j=j+1$
    	    \ENDIF
    	    \IF{$j > n_{tolerance}$}
    	        \RETURN $\boldsymbol{\theta}_{best}$
    	    \ENDIF
		\ENDFOR
        \RETURN $\boldsymbol{\theta}_{best}$
	\end{algorithmic}
\end{algorithm}

\section{Simulations}
\label{sec:sim}

In this section, we demonstrate the performance of the introduced Markovian RNN model both on real and synthetic datasets. We show the improvements achieved by our model by comparing the performance with vanilla RNN and commonly used, particularly in econometrics, conventional methods such as ARIMA, Markovian switching ARIMA (MS-ARIMA)~\cite{hamilton}, Kim-Nelson-Startz (KNS) model~\cite{kns}, and Filardo model with time-varying transition probabilities (TVTP)~\cite{tvtp}. In the first part, we simulate three synthetic sequences with two regimes and analyze the inference and switching capacity of our model under different scenarios. In the second set of experiments, we demonstrate the performance enhancements obtained by Markovian RNN in three real-life financial datasets and compare our results with the results of other methods. Also, we investigate the inferred regimes for these datasets by interpreting the temporal evolution of belief state vector and switching behavior of Markovian RNN.

\subsection{Synthetic Dataset Experiments}
\label{sec:sim_synth}

In order to analyze the capability of Markovian RNN to detect different regimes, and to investigate the switching behavior between these regimes, we conduct initial experiments on synthetic data. We first describe the simulation setups for synthetic data generation and then, present the results obtained by all methods on the synthetic datasets. We also investigate the effect of error covariance smoothing parameter ($\beta$) introduced in \eqref{eq:R_update} and mentioned in Remark \ref{remark:beta}.

\subsubsection{Simulation Setups}

In the synthetic data experiments, our goal is to predict the output $y_{t}$ given the input data $x_{1:t}$ such that $x_{t} \in \mathbb{R}$ is a scalar. The output data is given by $y_{t} = x_{t+1}$. Here, the goal of these experiments is to conceptually show the effectiveness of our algorithm. In order to demonstrate the learning behavior of our algorithm with different patterns and switching scenarios, simulated sequences should have various regimes, where each regime possesses different temporal statistics. To this end, we conduct three experiments in which we simulate autoregressive processes with deterministic and Markovian switching, and a sinusoidal with Markovian switching.

\begin{figure}
\centering
\begin{subfigure}[b]{0.49\textwidth}
   \includegraphics[width=0.98\linewidth]{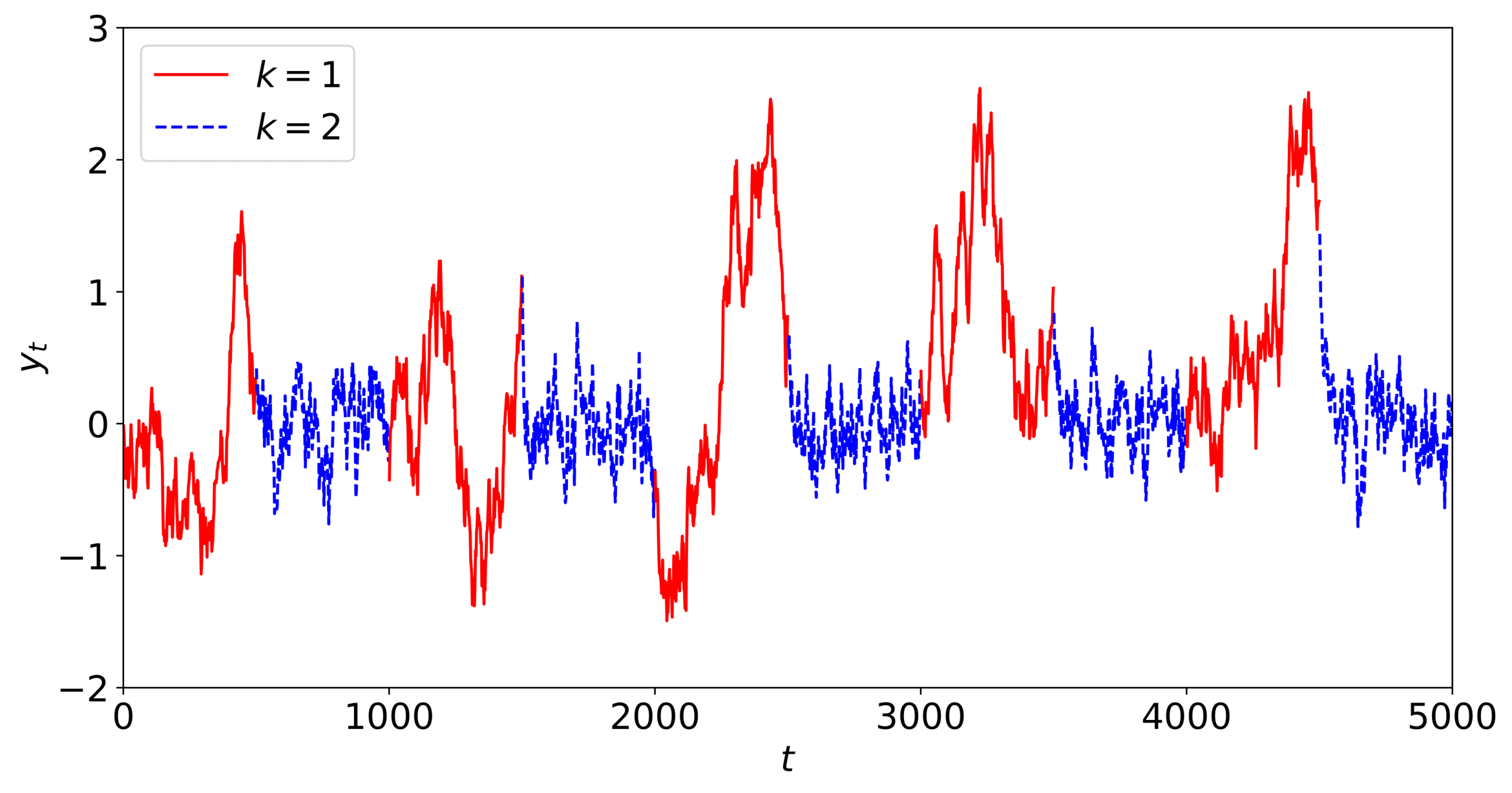}
   \caption{AR(1) process with two regimes and deterministic switching}
   \label{fig:synth_1} 
\end{subfigure}
\begin{subfigure}[b]{0.49\textwidth}
   \includegraphics[width=0.98\linewidth]{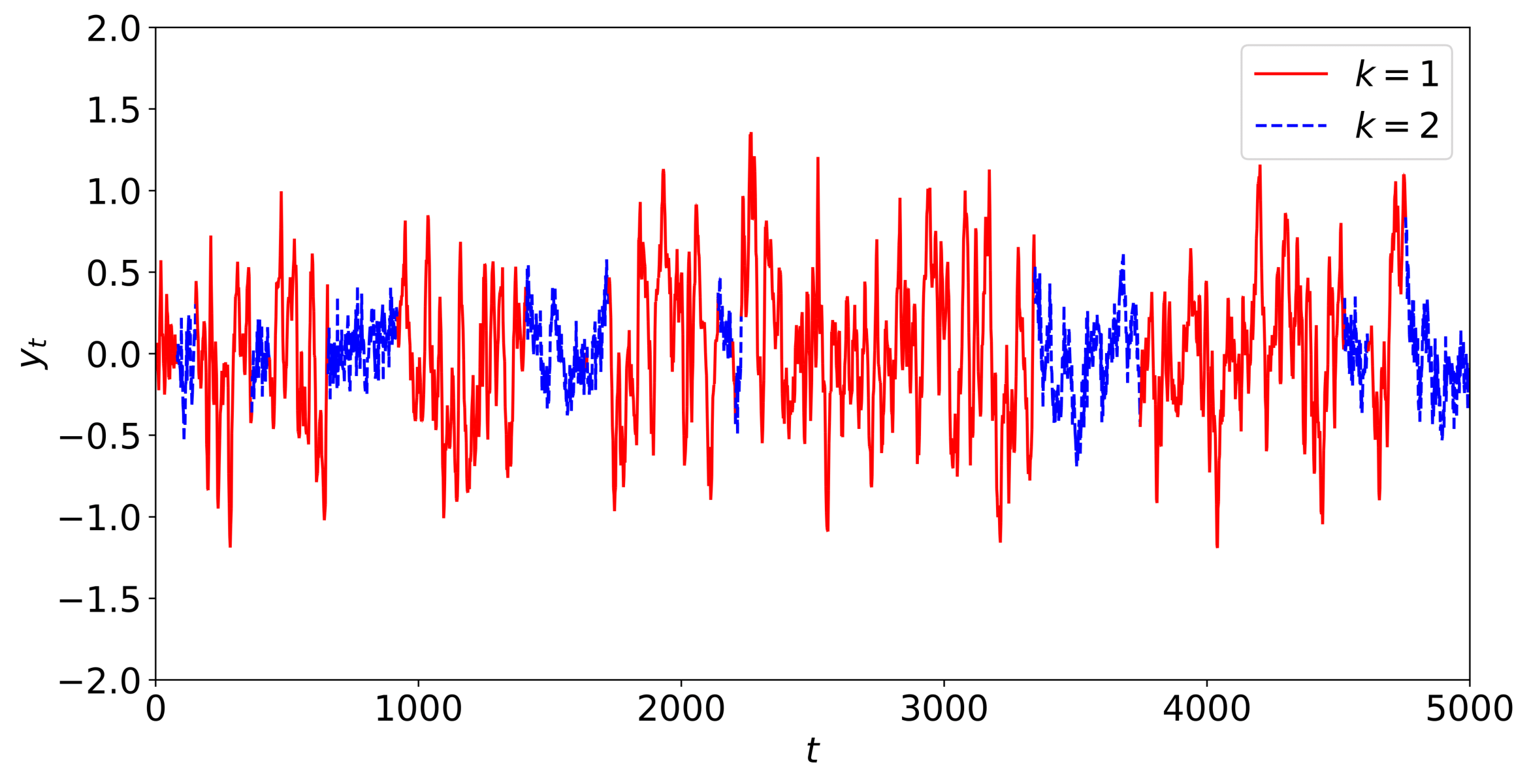}
   \caption{AR(3) process with two regimes and Markovian switching}
   \label{fig:synth_2}
\end{subfigure}
\begin{subfigure}[b]{0.49\textwidth}
   \includegraphics[width=0.98\linewidth]{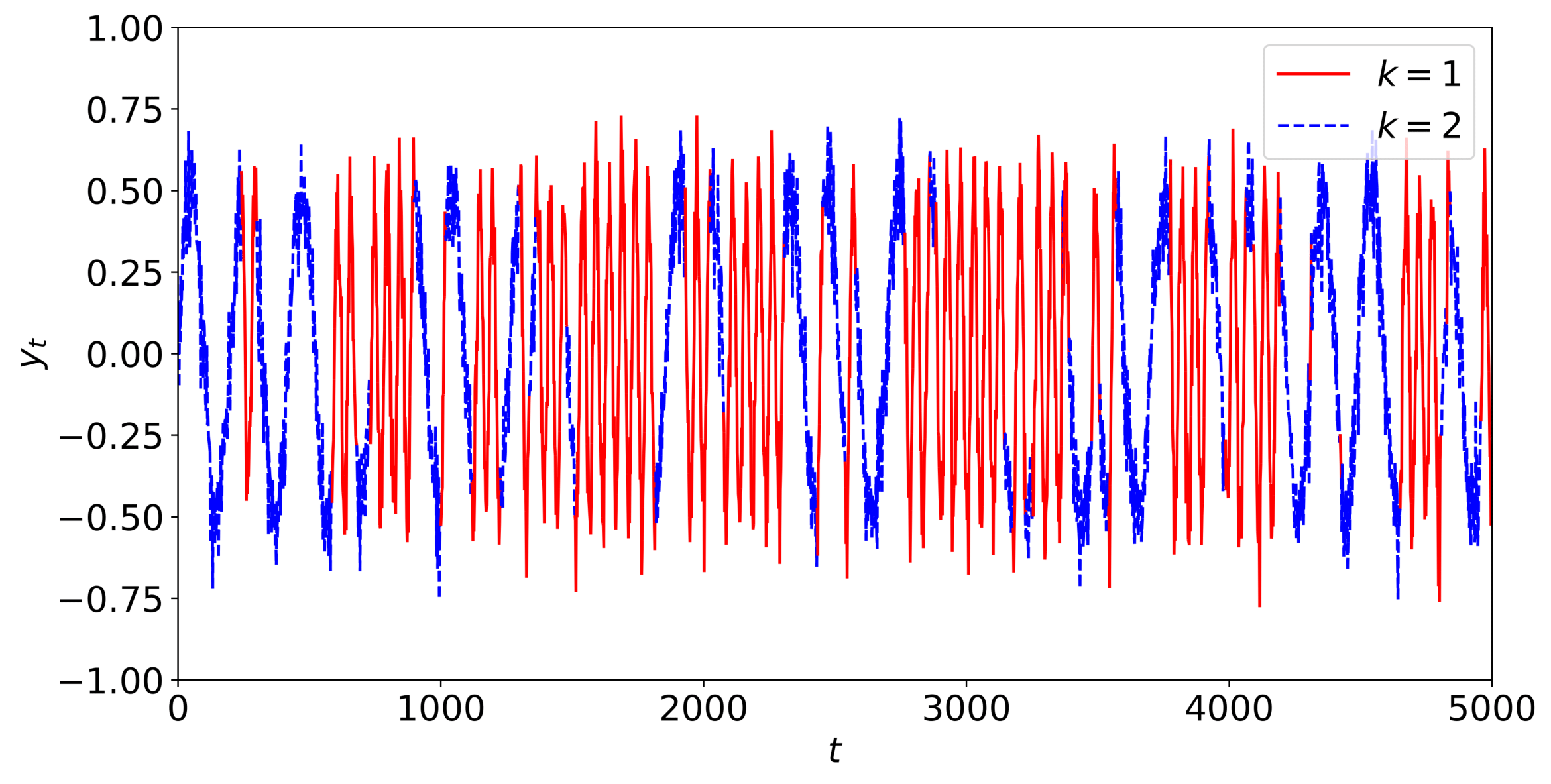}
   \caption{Sinusoidal with two regimes and Markovian switching}
   \label{fig:synth_3}
\end{subfigure}
\label{fig:synth}
\caption{Illustrations of simulated sequences for synthetic dataset experiments. Red color is used for the first regime and blue color is used for the second regime.}
\end{figure}

\paragraph{Autoregressive Process with Deterministic Switching}
In the first synthetic dataset experiment, we aim to generate a sequence with sharp transitions and obvious distinctions between regimes. To this end, we generate an autoregressive (AR) process with deterministic switching, which is given with the following set of equations:
\noindent
\begin{equation}
    x_{t+1} = \begin{cases}
            x_{t} + \epsilon \quad \text{if} \mod(t, 1000) < 500,  \\
            -0.9\,x_{t} + \epsilon \quad \text{if} \mod(t, 1000) \geq 500.
          \end{cases}
\label{eq:synth_data}
\end{equation}
where $x_t \in \mathbb{R}$ is the value of the time series at the $t^{th}$ time step, and $\epsilon \sim \mathcal{N}(0, 0.01)$ is the process noise. Here, \eqref{eq:synth_data} describes an AR process with two equal-duration (500) regimes, in which the system osciallates between. The first regime describes a random walk process, whereas the second regime gradually drifts towards white noise. The simulated system is deterministic in terms of switching mechanism between regimes since it strictly depends on the time step. Fig. \ref{fig:synth_1} demonstrates the time series generated by this setup.

\paragraph{Autoregressive Process with Markovian Switching}
In this setup, we consider Markovian switching instead of deterministic switching. Here, the transition between regimes has Markovian property, therefore the regime of next time step only depends on the current regime. We consider third order AR processes with the coefficients of $\{0.95, 0.5, -0.5\}$ and $\{0.95, -0.5, 0.5\}$ for each regime respectively, and $\epsilon \sim \mathcal{N}(0, 0.01)$. For the transition matrix, we consider $\mathbf{\Psi} = \big[\begin{smallmatrix} 0.998 & 0.002 \\ 0.004 & 0.996 \end{smallmatrix}\big]$. Fig. \ref{fig:synth_2} demonstrates the time series generated by this setup.

\paragraph{Sinusoidal with Markovian Switching}
In this experiment, we generate a noisy sinusoidal signal with two regimes, where every regime represents a different frequency. Here, the simulated signal has two regimes with the magnitude of $0.5$ and periods of $50$ and $200$ for the generated sinusoidals. The whole sequence consists of $5000$ time steps and Markovian switching is controlled by the transition matrix $\mathbf{\Psi} = \big[\begin{smallmatrix} 0.99 & 0.01 \\ 0.01 & 0.99 \end{smallmatrix}\big]$. We also scale the magnitude to half and add Gaussian noise to the generated signal ($\epsilon \sim \mathcal{N}(0, 0.0025)$). 

\subsubsection{Synthetic Dataset Performance}
Here, we present the training procedure and the results of the methods in terms of the root mean squared error (RMSE) and mean absolute error (MAE). In these experiments, each synthetic time series has 5000 time steps of temporal length. We split the data into three splits for training (60$\%$), validation (20$\%$), and test (20$\%$) respectively. We perform training on the training set and choose the best configuration based on the performance in the validation set. Then, we compare the test results of the best configuration for each method.

For the parameter search of vanilla RNN and Markovian RNN, we perform grid search for the number of hidden dimensions in the range of $n_h = \{4, \ldots, 64\}$, truncation length in $\tau = \{2, \ldots, 64\}$, and learning rate in $\eta = [0.00001, 0.03]$. We employ Xavier initialization for weight initialization. We initialize the rows of the transition matrix ($\mathbf{\Psi}$) in Markovian RNN using the Dirichlet distribution with the concentration vector $\boldsymbol{\rho}^{(k)}$ such that $\rho_i^{(k)} = \rho_0$ if $i=k$ and $\frac{1-\rho_0}{K-1}$ otherwise, where $k$ is the regime index. Here, $\rho_0$ determines the concentration over the diagonal elements of the initialized transition matrix. We search $\rho_0$ and $\beta$ on the interval of $[0.3, 0.95]$. For the number of internal regimes in Markovian RNN, we search on the interval $K=\{2, \ldots, 5\}$. We also perform early stopping based on validation error such that we stop the training if the loss does not decrease for $20$ consecutive epochs or the number of epochs reaches to $200$. For other methods, we determine the component order ranges through analying autocorrelation and partial autocorrelation plots, and use the parameters that result with the best validation performance. We obtained the best validation results for Markovian RNN in these experiments using the following parameters respectively: $n_h = \{16, 8, 16\}$, $\rho_0 = \{0.5, 0.6, 0.7\}$, $\beta = \{0.7, 0.5, 0.9\}$, $\tau = \{4, 8, 8\}$, $K = \{2, 2, 2\}$, $\eta = \{0.0003, 0.0001, 0.003\}$.

\renewcommand{\tabcolsep}{5pt}
\begin{table}[t]
\centering
\begin{tabular}{|c|c|c|c|c|c|c|}
\hline
\multicolumn{1}{|l|}{\multirow{3}{*}{}} & \multicolumn{6}{c|}{Simulation Setup} \\ \cline{2-7} 
\multicolumn{1}{|l|}{} & \multicolumn{2}{c|}{AR (det.)} & \multicolumn{2}{c|}{AR (mar.)} & \multicolumn{2}{c|}{Sinusoidal (mar.)} \\ \cline{2-7} 
\multicolumn{1}{|l|}{} & RMSE & MAE & RMSE & MAE & RMSE & MAE \\ \hline
ARIMA & 0.333 & 0.228 & 0.183 & 0.148 & 0.136 & 0.108 \\ \hline
MS-ARIMA~\cite{hamilton} & 0.206 & 0.145 & 0.148 & 0.120 & 0.128 & 0.103 \\ \hline
KNS~\cite{kns} & 0.447 & 0.271 & 0.196 & 0.155 & 0.142 & 0.114 \\ \hline
TVTP~\cite{tvtp} & 0.206 & 0.145 & 0.160 & 0.129 & 0.136 & 0.108 \\ \hline
Vanilla RNN~\cite{Elman90} & 0.193 & 0.134 & 0.146 & 0.113 & 0.126 & 0.099 \\ \hline
\textbf{Markovian RNN} & \textbf{0.178} & \textbf{0.120} & \textbf{0.126} & \textbf{0.097} & \textbf{0.121} & \textbf{0.091} \\ \hline
\end{tabular}
\caption{Synthetic dataset experiment results for baseline methods and the introduced Markovian RNN in terms of RMSE and MAE.}
\label{tab:result_synth}
\end{table}

In Table \ref{tab:result_synth}, we provide the test RMSE and MAE obtained by each method on the synthetic datasets. In all setups, our model performs significantly better than other methods. Regardless of the switching mechanism and process dynamics, our model brings considerable improvements in terms of prediction RMSE and MAE. The performance Kim, Nelson and Startz model is not competitive with other methods, since it relies on switching variance between regimes. TVTP also has the form of MS-ARIMA, but it also models the transition probabilities as temporally varying values. In our setups, the transition probabilities are fixed, hence this method does not bring any improvement over MS-ARIMA. MS-ARIMA works significantly more accurately than standard ARIMA as expected. Likewise, Markovian RNN enjoys the benefits of adaptive HMM-based switching, which improves the predictions compared to the predictions of vanilla RNN.

\renewcommand{\tabcolsep}{2.5pt}
\begin{table}[t]
\centering
\begin{tabular}{|c|c|c|c|c|c|c|}
\hline
\multicolumn{1}{|l|}{\multirow{3}{*}{}} & \multicolumn{6}{c|}{Dataset} \\ \cline{2-7} 
\multicolumn{1}{|l|}{} & \multicolumn{2}{c|}{\textit{USD/EUR}} & \multicolumn{2}{c|}{\textit{USD/GBP}} & \multicolumn{2}{c|}{\textit{USD/TRY}} \\ \cline{2-7} 
\multicolumn{1}{|l|}{} & RMSE & MAE & RMSE & MAE & RMSE & MAE \\ \hline
ARIMA & 2.34E-3 & 1.78E-3 & 2.69E-3 & 1.98E-3 & 5.42E-2 & 2.67E-2 \\ \hline
MS-ARIMA~\cite{hamilton} & 2.28E-3 & 1.73E-3 & 2.53E-3 & 1.88E-3 & 5.34E-2 & 2.63E-2 \\ \hline
KNS~\cite{kns} & 2.14E-3 & 1.54E-3 & 2.61E-3 & 1.85E-3 & 5.40E-2 & 2.57E-2 \\ \hline
TVTP~\cite{tvtp} & 2.37E-3 & 1.81E-3 & 2.47E-3 & 1.79E-3 & 5.29E-2 & 2.50E-2 \\ \hline
Vanilla RNN~\cite{Elman90} & 2.24E-3 & 1.68E-3 & 2.60E-3 & 1.84E-3 & 5.36E-2 & 2.57E-2 \\ \hline
\textbf{Markovian RNN} & \textbf{2.02E-3} & \textbf{1.51E-3} & \textbf{2.34E-3} & \textbf{1.69E-3} & \textbf{5.10E-2} & \textbf{2.34E-2} \\ \hline
\end{tabular}
\caption{\textit{USD/EUR}, \textit{USD/GBP} and \textit{USD/TRY} dataset experiment results for baseline methods and the introduced Markovian RNN in terms of RMSE and MAE.}
\label{tab:result_real}
\end{table}

\begin{figure*}[h!]
\centering
\begin{subfigure}[t]{0.47\textwidth}
   \centering
\includegraphics[width=0.94\textwidth]{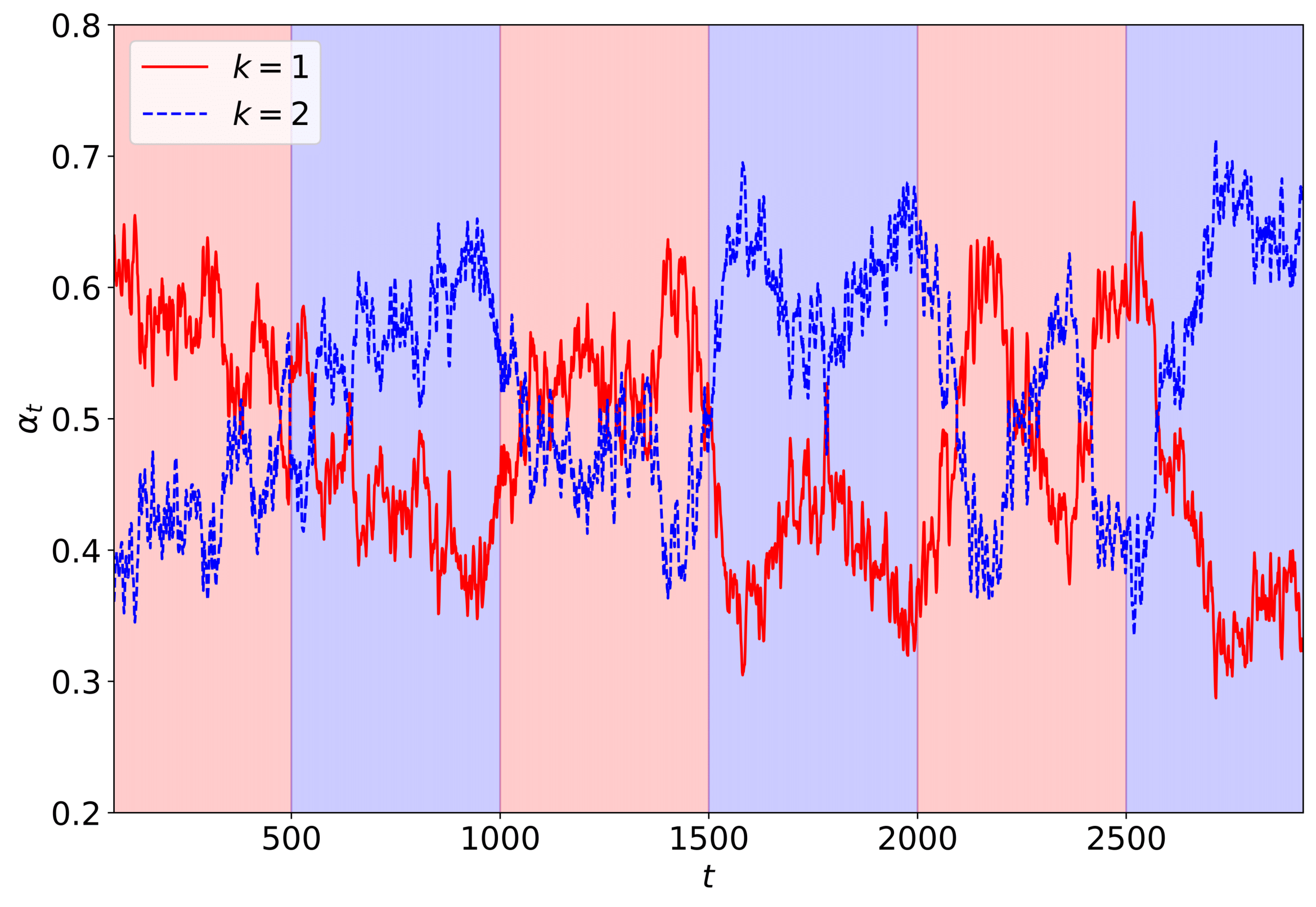}
\caption{Belief state vector of Markovian RNN for AR process sequence with deterministic switching} \label{fig:results_bs1}
\end{subfigure}
\hfill
 \begin{subfigure}[t]{0.47\textwidth}
   \centering
\includegraphics[width=0.94\textwidth]{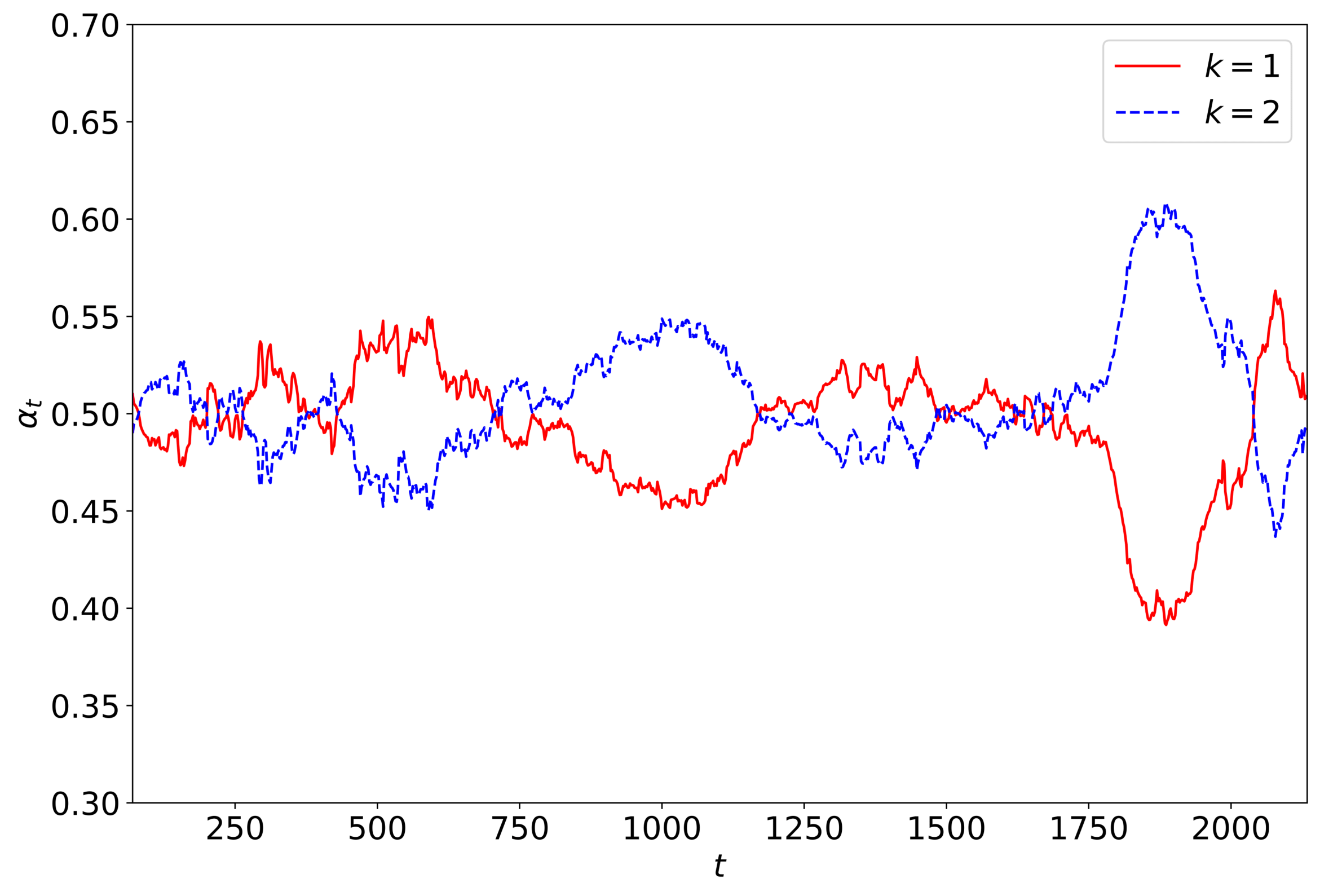}
\caption{Belief state vector of Markovian RNN for \textit{USD/EUR} dataset} \label{fig:results_bs2}
\end{subfigure}
\caption{Belief state vectors of Markovian RNN for two experiments. In Fig. \ref{fig:results_bs1}, background colors represent the real regime value, where red color is used for the first regime and blue color is used for the second regime. Our model can properly distinguish between the two regimes, thus the resulting belief state vector values are consistent with the real regimes. In Fig. \ref{fig:results_bs2}, since the experiment is performed on a real-life dataset, in which the real regime values are not observable, consistency analysis is not possible. However, we still observe that our model switches between regimes in a stable way without any saturation.}\label{fig:results_bs}.
\end{figure*}

\begin{figure}[h!]
\centering
\begin{subfigure}[b]{0.45\textwidth}
\centering
    \includegraphics[width=0.9\columnwidth]{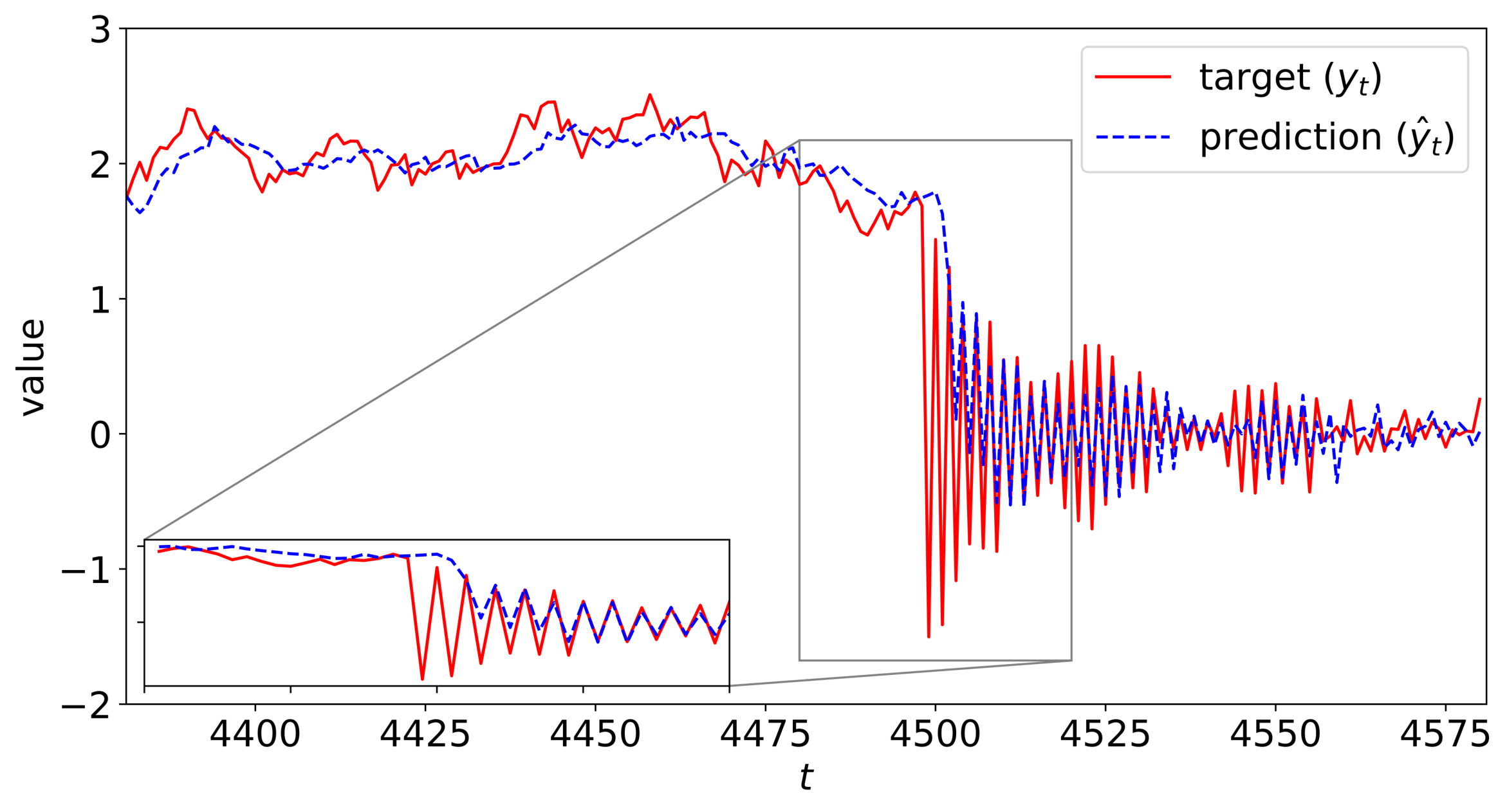}
    \caption{$\beta=0.5$}
    \label{fig:zoom_1}
\end{subfigure}
\begin{subfigure}[b]{0.45\textwidth}
\centering
    \includegraphics[width=0.9\columnwidth]{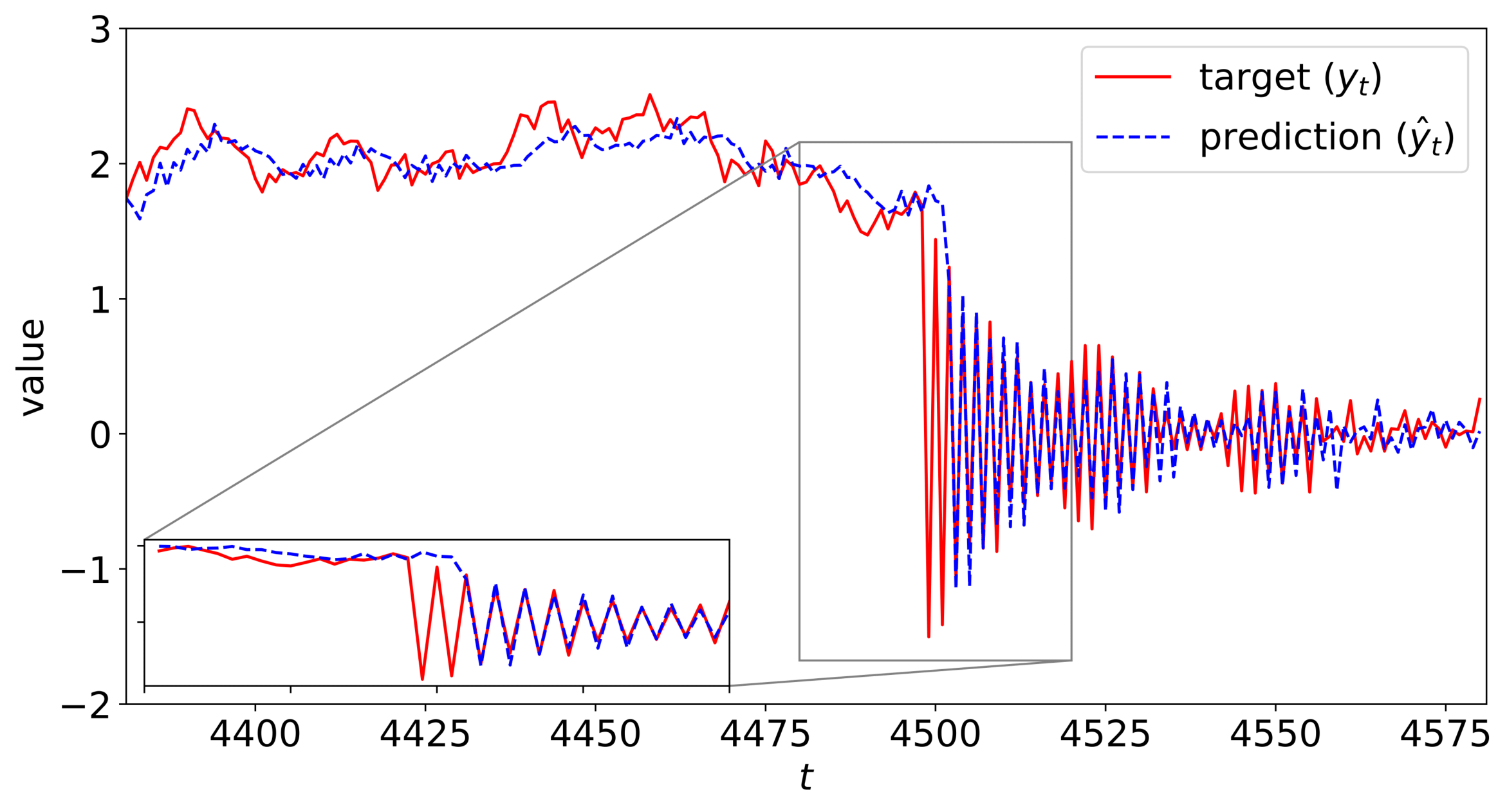}
    \caption{$\beta=0.7$}
    \label{fig:zoom_2}
\end{subfigure}
\begin{subfigure}[b]{0.45\textwidth}
\centering
    \includegraphics[width=0.9\columnwidth]{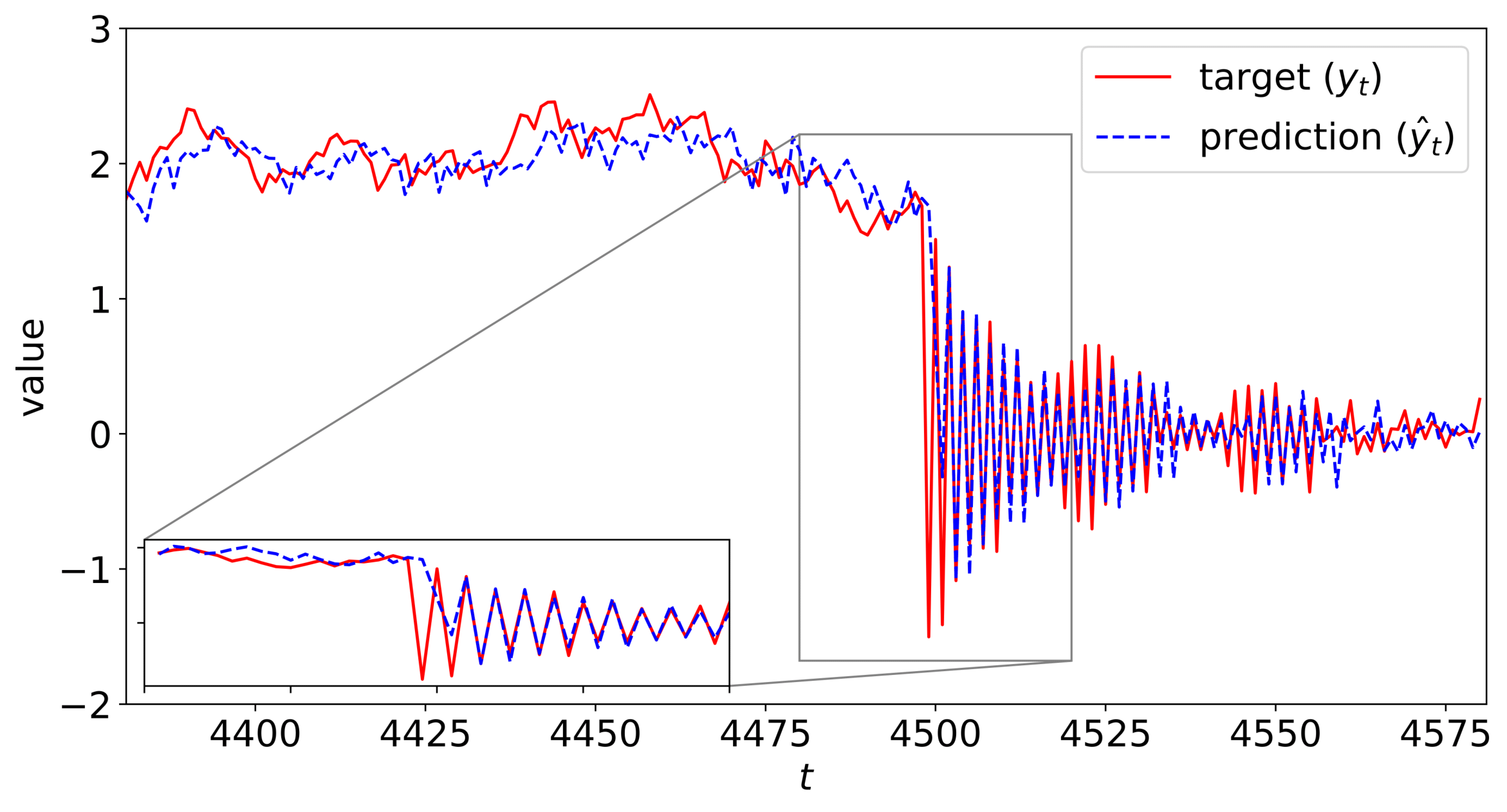}
    \caption{$\beta=0.9$}
    \label{fig:zoom_3}
\end{subfigure}
\label{fig:zoom}
\caption{Markovian RNN predictions on the test set of the AR process with deterministic switching, and the zoomed in plot at the regime switching region for different error covariance smoothing parameters.}
\end{figure}

\subsection{Real Dataset Experiments}

\label{sec:sim_real}

In this section, we provide the performance of our model and other methods in two financial datasets. We consider \textit{USD/EUR}, \textit{USD/GBP} and \textit{USD/TRY} currency ratio data~\cite{data} from January 1, 2010 to January 24, 2020 with hourly resolution for these experiments. We calculate the mean and variance of prices for every day, and work in daily resolution. The goal is to predict the average currency rate for the $t^{th}$ day ($y_{t}$) given the input data $\boldsymbol{x}_{1:t}$. In this case, the input data is a two-dimensional vector such that $\boldsymbol{x}_{t} \in \mathbb{R}^2$ since we consider $\boldsymbol{x}_t$ to include the mean and standard deviation of the currency rate for the $t^{th}$ day. We split the data into three splits for training (60$\%$), validation (20$\%$), and test (20$\%$) respectively. 

For the parameter search of vanilla RNN and Markovian RNN, we perform grid search for the number of hidden dimensions in the range of $n_h = \{4, \ldots, 64\}$, truncation length in $\tau = \{2, \ldots, 64\}$, and learning rate in $\eta = [0.00001, 0.1]$. We employ Xavier initialization for weight initialization. We search $\rho_0$ and $\beta$ on the interval of $[0.3, 0.95]$. For the number of internal regimes in Markovian RNN, we search on the interval $K=\{2, \ldots, 5\}$. We take the first-order difference of the data to decrease the trend effect. Before testing, we calibrate the predictions by fitting a linear regressor on validation set, which minimizes the residual sum of least squares between validation predictions and validation labels. We also perform early stopping based on validation error such that we stop the training if the loss does not decrease for $20$ consecutive epochs or the number of epochs reaches to $200$. We obtained the best validation results for Markovian RNN in these experiments using the following parameters respectively: $n_h = \{32, 32, 64\}$, $\rho_0 = \{0.7, 0.5, 0.7\}$, $\beta = \{0.9, 0.5, 0.7\}$, $\tau = \{32, 16, 64\}$, $K = \{2, 3, 3\}$, $\eta = \{0.003, 0.001, 0.0003\}$.

In Table \ref{tab:result_real}, we provide the test RMSE and MAE values obtained by each method on these currency datasets. All methods perform worst in \textit{USD/TRY} due to the high oscillations in the sequence. In all cases, Markovian RNN performs significantly better than other methods. Our model obtains the lowest RMSE and MAE values thanks to the efficient combination of nonlinear modeling capability of RNNs and adaptive switching controlled by HMM. Although the regimes that affect the sequential dynamics cannot be observed, our model is able to detect these regions and switch between them to adjust the predictions for nonstationarity. We further analyze the switching behavior of Markovian RNN in Section \ref{sec:sim_analysis} and investigate the inferred regime belief vectors.

\subsection{Regime Switching Behavior of Markovian RNN}

\label{sec:sim_analysis}

To investigate how our model detects the regimes and decides to switch from one to another, we illustrate the belief state vector through time in Fig. \ref{fig:results_bs1} and \ref{fig:results_bs2} for AR process sequence with deterministic switching and \textit{USD/EUR} dataset, respectively. For instance, in Fig. \ref{fig:results_bs1}, we observe that our model can determine different regimes correctly and adjust the belief vector accordingly. Here, background colors are used to indicate the real regime. During the analysis of real-life dataset experiments, checking the consistency between belief state vector and real regimes is not possible, since they are not observable. In Fig. \ref{fig:results_bs2}, we observe that there are certain periods in which the second regime dominates the predictions. Considering the switching behavior in Fig. \ref{fig:results_bs2}, we can interpret that our model does not suffer from rapid oscillations between regimes or does not saturate at a regime. The evolution of the belief state vector is stable, but still responsive to nonstaionarity.

In addition, we zoom in to the predictions of Markovian RNN around switching regions in Fig. \ref{fig:zoom_1}, and show that our model can successfully adapt to nonstationarity. As we mention in Remark \ref{remark:beta}, the error covariance smoothing parameter ($\beta$) in \eqref{eq:R_update} can be tuned with cross-validation to adjust the switching behavior of Markovian RNN. For instance, lower values of this parameter ($\beta=0.5$) bring more toleration for prediction errors and may provide robustness against outliers with the expense of lagged transitions between regimes as shown in Fig. \ref{fig:zoom_1}. On the contrary, higher values ($\beta=0.9$) provide faster transitions but may cause oscillating predictions or lead to instability.

\section{Conclusion}\label{sec:concl}

We study nonlinear regression for time series prediction in nonstationary environments. We introduce a novel time series prediction network, Markovian RNN, which is an RNN with multiple internal regimes and, HMM-based switching. Each internal regime controls the hidden state transitions with different weights. We employ an HMM to control the switching mechanism between the internal regimes, and jointly optimize the whole network in an end-to-end fashion. By combining the nonlinear representation capability of RNNs and the adaptivity obtained thanks to HMM-based switching, our model can capture nonlinear temporal patterns in highly nonstationary environments, in which the underlying system that generates the time series has temporally varying dynamics.

Through an extensive set of synthetic and real-life dataset experiments, we demonstrate the performance gains compared to the conventional methods such as vanilla RNN~\cite{Elman90}, MS-ARIMA~\cite{hamilton} and Filardo model~\cite{tvtp}, which are commonly preferred in business, economy and finance applications. We show that the introduced model performs significantly better than other methods in terms of prediction RMSE and MAE thanks to the joint optimization and the efficient combination of nonlinear regression with RNNs, and HMM-based regime switching. We show that Markovian RNN can properly determine the regimes and switch between them to make more accurate predictions. We also analyze the effect of the error covariance smoothing parameter on the switching behavior of our model. As the experimental results and our analysis indicate, our model can capture nonlinear temporal patterns while successfully adapting nonstationarity without any instability or saturation issues.

\appendices

\section{}
\label{appendix}

In this part, we provide the derivatives of the model weights ($\boldsymbol{\theta} = \{\{\mathbf{W}_{xh}\}_{k=1}^K, \{\{\mathbf{W}_{hh}\}_{k=1}^K, \mathbf{W}_{hy}, \mathbf{\Psi}\}$) of Markovian RNN. The equations of the basic derivatives are as follows:
\begin{align}
    &\frac{\partial \ell_t}{\partial \boldsymbol{\hat{y}}_t} = -2\boldsymbol{e}_t^T, \label{eq:ly} \\
    &\frac{\partial \boldsymbol{\hat{y}}_t}{\partial \boldsymbol{h}_t} = \mathbf{W}_{hy}, \label{eq:yh}\\
    &\frac{\partial \boldsymbol{h}_t}{\partial \boldsymbol{h}_t^{(k)}} = \alpha_{t-1, k}, \label{eq:hh}\\
    &\frac{\partial \boldsymbol{h}_t^{(k)}}{\partial \boldsymbol{h}_{t-1}} = \mathbf{W}_{hh}^{(k)} \odot \mathrm{diag}({f_h^{'}(\boldsymbol{z}_{t}^{(k)})}), \label{eq:hh_}\\
    &\frac{\partial \boldsymbol{h}_t}{\partial \alpha_{t-1, k}} = \boldsymbol{h}_t^{(k)}, \label{eq:ha}\\
    &\frac{\partial \alpha_{t, k}}{\partial \Tilde{\alpha}_{t, k'}} = \frac{\delta_{kk'}\text{sum}( \boldsymbol{\Tilde{\alpha}}_t)-\Tilde{\alpha}_{t, k}}{\text{sum}( \boldsymbol{\Tilde{\alpha}}_t)^2}, \label{eq:aa}
\end{align}
where $\boldsymbol{z}_t^{(k)} = \mathbf{W}^{(k)}_{hh} \boldsymbol{h}_{t-1} + \mathbf{W}^{(k)}_{xh} \boldsymbol{x}_{t}$. We use \eqref{eq:hh} and \eqref{eq:hh_} to obtain the following:
\begin{align}
    &\frac{\partial \boldsymbol{h}_t}{\partial \boldsymbol{h}_{t-\tau}} = \prod\limits_{t'=t-\tau+1}^{t}\frac{\partial \boldsymbol{h}_t'}{\partial \boldsymbol{h}_{t'-1}}, \label{eq:hh__}
\end{align}
where $\frac{\partial \boldsymbol{h}_t}{\partial \boldsymbol{h}_{t-1}} = \sum\limits_{k=1}^K \alpha_{t-1, k} \mathbf{W}_{hh}^{(k)} \odot \mathrm{diag}({f_h^{'}(\boldsymbol{z}_{t}^{(k)})})$. Then, we can use \eqref{eq:ly}-\eqref{eq:hh_} and \eqref{eq:hh__} to obtain $\frac{\partial \ell_t}{\partial \mathbf{W}_{xh}^{(k)}}$ and $\frac{\partial \ell_t}{\partial \mathbf{W}_{hh}^{(k)}}$ as follows:
\begin{align}
    &\frac{\partial \ell_t}{\partial \mathbf{W}_{xh}^{(k)}} = \sum\limits_{t'=t-\tau}^{t}
    \frac{\partial \ell_{t}}{\partial \boldsymbol{\hat{y}}_{t}} 
    \frac{\partial \boldsymbol{\hat{y}}_{t}}{\partial \boldsymbol{h}_{t}} 
    \frac{\partial \boldsymbol{h}_{t}}{\partial \boldsymbol{h}_{t'}}
    \frac{\partial \boldsymbol{h}_{t'}}{\partial \boldsymbol{h}_{t'}^{(k)}}
    \frac{\partial \boldsymbol{h}_{t'}^{(k)}}{\partial \mathbf{W}_{xh}^{(k)}}, \label{eq:wxh} \\ 
    &\frac{\partial \ell_t}{\partial \mathbf{W}_{hh}^{(k)}} = \sum\limits_{t'=t-\tau}^{t}
    \frac{\partial \ell_{t}}{\partial \boldsymbol{\hat{y}}_{t}} 
    \frac{\partial \boldsymbol{\hat{y}}_{t}}{\partial \boldsymbol{h}_{t}} 
    \frac{\partial \boldsymbol{h}_{t}}{\partial \boldsymbol{h}_{t'}}
    \frac{\partial \boldsymbol{h}_{t'}}{\partial \boldsymbol{h}_{t'}^{(k)}}
    \frac{\partial \boldsymbol{h}_{t'}^{(k)}}{\partial \mathbf{W}_{hh}^{(k)}},\label{eq:whh}
\end{align}
where $\frac{\partial \boldsymbol{h}_{t'}^{(k)}}{\partial \mathbf{W}_{xh}^{(k)}} = \begin{bmatrix}
        \frac{\partial \boldsymbol{h}_{t'}^{(k)}}{\partial {W_{xh,ij}^{(k)}}} \\
    \end{bmatrix}$ such that $\frac{\partial \boldsymbol{h}_{t'}^{(k)}}{\partial {W_{xh,ij}^{(k)}}} = x_{t',j} \boldsymbol{d}_i \odot f_h'(\boldsymbol{z}_{t'}^{(k)})$, $\frac{\partial \boldsymbol{h}_{t'}^{(k)}}{\partial \mathbf{W}_{hh}^{(k)}} = \begin{bmatrix}
        \frac{\partial \boldsymbol{h}_{t'}^{(k)}}{\partial W_{hh,ij}^{(k)}} \\
    \end{bmatrix}$ such that $\frac{\partial \boldsymbol{h}_{t'}^{(k)}}{\partial W_{hh,ij}^{(k)}} = h_{t'-1,j} \boldsymbol{d}_i \odot f_h'(\boldsymbol{z}_{t'}^{(k)})$, and $\tau$ is the truncation length. Here, $\boldsymbol{d}$ is a vector such that $d_{i'} = \delta_{ii'}$.

Using \eqref{eq:ly}, we can calculate $\frac{\partial \ell_t}{\partial \mathbf{W}_{hy}}$ as:
\begin{align}
    &\frac{\partial \ell_t}{\partial \mathbf{W}_{hy}} = \frac{\partial \ell_{t}}{\partial \boldsymbol{\hat{y}}_{t}} \frac{\partial \boldsymbol{\hat{y}}_{t}}{\partial \mathbf{W}_{hy}} = -2 \boldsymbol{e}_{t}\boldsymbol{h}_{t}^T. \label{eq:why}
\end{align}

Finally, we can compute the derivative of the transition matrix $\mathbf{\mathbf{\Psi}}$ using \eqref{eq:ly}, \eqref{eq:yh}, \eqref{eq:ha} and \eqref{eq:hh__} as follows:
\begin{align}
    &\frac{\partial \ell_t}{\partial \mathbf{\Psi}} = \sum\limits_{t'=t-\tau}^{t} 
    \frac{\partial \ell_t}{\partial \boldsymbol{\hat{y}}_t} 
    \frac{\partial \boldsymbol{\hat{y}}_{t}}{\partial \boldsymbol{h}_{t}} 
    \frac{\partial \boldsymbol{h}_{t}}{\partial \boldsymbol{h}_{t'}}
    \frac{\partial \boldsymbol{h}_{t'}}{\partial \mathbf{\Psi}},
\end{align}
where $\frac{\partial \boldsymbol{h}_{t'}}{\partial \mathbf{\Psi}} = \sum\limits_{k=1}^K \boldsymbol{h}_{t'}^{(k)} \frac{\partial \alpha_{t'-1, k}}{\partial \mathbf{\Psi}}$. We can express the derivative terms in this summation as:
\begin{align}
   \frac{\partial \alpha_{t'-1, k}}{\partial \mathbf{\Psi}} &= \sum\limits_{k'=1}^K \frac{\partial \alpha_{t'-1, k}}{\partial \Tilde{\alpha}_{t'-1, k'}} \frac{\partial \Tilde{\alpha}_{t'-1, k'}}{\partial \mathbf{\Psi}},
\end{align}
where $\frac{\partial \Tilde{\alpha}_{t'-1, k'}}{\partial \mathbf{\Psi}} = \begin{bmatrix}
        \frac{\partial \Tilde{\alpha}_{t'-1, k'}}{\partial \Psi_{ij}} \\
    \end{bmatrix}$ such that $\frac{\partial \Tilde{\alpha}_{t'-1, k'}}{\partial \Psi_{ij}} = \delta_{i,k'}\phi_{t-1,k'}\alpha_{t'-2, j}$, and $\frac{\partial \alpha_{t'-1, k}}{\partial \Tilde{\alpha}_{t'-1, k'}}$ is given in \eqref{eq:aa}.

\bibliographystyle{IEEEtran}
\balance
\bibliography{main}

\begin{thebibliography}{10}
\providecommand{\url}[1]{#1}
\csname url@samestyle\endcsname
\providecommand{\newblock}{\relax}
\providecommand{\bibinfo}[2]{#2}
\providecommand{\BIBentrySTDinterwordspacing}{\spaceskip=0pt\relax}
\providecommand{\BIBentryALTinterwordstretchfactor}{4}
\providecommand{\BIBentryALTinterwordspacing}{\spaceskip=\fontdimen2\font plus
\BIBentryALTinterwordstretchfactor\fontdimen3\font minus
  \fontdimen4\font\relax}
\providecommand{\BIBforeignlanguage}[2]{{%
\expandafter\ifx\csname l@#1\endcsname\relax
\typeout{** WARNING: IEEEtran.bst: No hyphenation pattern has been}%
\typeout{** loaded for the language `#1'. Using the pattern for}%
\typeout{** the default language instead.}%
\else
\language=\csname l@#1\endcsname
\fi
#2}}
\providecommand{\BIBdecl}{\relax}
\BIBdecl

\bibitem{Haykin}
S.~Haykin, \emph{Neural Networks: A Comprehensive Foundation}, 2nd~ed.\hskip
  1em plus 0.5em minus 0.4em\relax Upper Saddle River, NJ, USA: Prentice Hall
  PTR, 1998.

\bibitem{Tolga2018}
T.~{Ergen} and S.~S. {Kozat}, ``Efficient online learning algorithms based on
  lstm neural networks,'' \emph{IEEE Transactions on Neural Networks and
  Learning Systems}, vol.~29, no.~8, pp. 3772--3783, Aug 2018.

\bibitem{lstm_}
K.~{Greff}, R.~K. {Srivastava}, J.~{Koutník}, B.~R. {Steunebrink}, and
  J.~{Schmidhuber}, ``Lstm: A search space odyssey,'' \emph{IEEE Transactions
  on Neural Networks and Learning Systems}, vol.~28, no.~10, pp. 2222--2232,
  2017.

\bibitem{CBianchi2006}
N.~Cesa-Bianchi and G.~Lugosi, \emph{Prediction, Learning, and Games}.\hskip
  1em plus 0.5em minus 0.4em\relax New York, NY, USA: Cambridge University
  Press, 2006.

\bibitem{Vovk1990}
V.~G. Vovk, ``Aggregating strategies,'' in \emph{Proceedings of the Third
  Annual Workshop on Computational Learning Theory}, ser. COLT '90.\hskip 1em
  plus 0.5em minus 0.4em\relax San Francisco, CA, USA: Morgan Kaufmann
  Publishers Inc., 1990, pp. 371--386.

\bibitem{Yuksel}
S.~E. {Yuksel}, J.~N. {Wilson}, and P.~D. {Gader}, ``Twenty years of mixture of
  experts,'' \emph{IEEE Transactions on Neural Networks and Learning Systems},
  vol.~23, no.~8, pp. 1177--1193, 2012.

\bibitem{Singer94}
A.~C. Singer, G.~W. Wornell, and A.~V. Oppenheim, ``Nonlinear autoregressive
  modeling and estimation in the presence of noise,'' \emph{Digital Signal
  Processing}, vol.~4, no.~4, pp. 207 -- 221, 1994.

\bibitem{Engel2004}
Y.~{Engel}, S.~{Mannor}, and R.~{Meir}, ``The kernel recursive least-squares
  algorithm,'' \emph{IEEE Transactions on Signal Processing}, vol.~52, no.~8,
  pp. 2275--2285, Aug 2004.

\bibitem{zhang}
L.~{Zhang}, Y.~{Zhu}, and W.~X. {Zheng}, ``Energy-to-peak state estimation for
  markov jump rnns with time-varying delays via nonsynchronous filter with
  nonstationary mode transitions,'' \emph{IEEE Transactions on Neural Networks
  and Learning Systems}, vol.~26, no.~10, pp. 2346--2356, 2015.

\bibitem{Denizcan2017}
N.~D. {Vanli}, M.~O. {Sayin}, I.~{Delibalta}, and S.~S. {Kozat}, ``Sequential
  nonlinear learning for distributed multiagent systems via extreme learning
  machines,'' \emph{IEEE Transactions on Neural Networks and Learning Systems},
  vol.~28, no.~3, pp. 546--558, March 2017.

\bibitem{Raginsky}
M.~{Raginsky}, R.~M. {Willett}, C.~{Horn}, J.~{Silva}, and R.~F. {Marcia},
  ``Sequential anomaly detection in the presence of noise and limited
  feedback,'' \emph{IEEE Transactions on Information Theory}, vol.~58, no.~8,
  pp. 5544--5562, 2012.

\bibitem{miranian}
A.~{Miranian} and M.~{Abdollahzade}, ``Developing a local least-squares support
  vector machines-based neuro-fuzzy model for nonlinear and chaotic time series
  prediction,'' \emph{IEEE Transactions on Neural Networks and Learning
  Systems}, vol.~24, no.~2, pp. 207--218, 2013.

\bibitem{Shao14}
L.~{Shao}, D.~{Wu}, and X.~{Li}, ``Learning deep and wide: A spectral method
  for learning deep networks,'' \emph{IEEE Transactions on Neural Networks and
  Learning Systems}, vol.~25, no.~12, pp. 2303--2308, Dec 2014.

\bibitem{ObjDet}
Z.~{Zhao}, P.~{Zheng}, S.~{Xu}, and X.~{Wu}, ``Object detection with deep
  learning: A review,'' \emph{IEEE Transactions on Neural Networks and Learning
  Systems}, vol.~30, no.~11, pp. 3212--3232, 2019.

\bibitem{DeepRec13}
M.~Hermans and B.~Schrauwen, ``Training and analysing deep recurrent neural
  networks,'' in \emph{Advances in Neural Information Processing Systems 26},
  C.~J.~C. Burges, L.~Bottou, M.~Welling, Z.~Ghahramani, and K.~Q. Weinberger,
  Eds.\hskip 1em plus 0.5em minus 0.4em\relax Curran Associates, Inc., 2013,
  pp. 190--198.

\bibitem{Liu16}
P.~Liu, X.~Qiu, and X.~Huang, ``Recurrent neural network for text
  classification with multi-task learning,'' 2016.

\bibitem{Laptev2017}
N.~Laptev, J.~Yosinski, L.~E. Li, and S.~Smyl, ``Time-series extreme event
  forecasting with neural networks at uber,'' in \emph{International Conference
  on Machine Learning}, no.~34, 2017, pp. 1--5.

\bibitem{DeMulder2015}
W.~D. Mulder, S.~Bethard, and M.-F. Moens, ``A survey on the application of
  recurrent neural networks to statistical language modeling,'' \emph{Computer
  Speech \& Language}, vol.~30, no.~1, pp. 61 -- 98, 2015.

\bibitem{Lin96}
{Tsungnan Lin}, B.~G. {Horne}, P.~{Tino}, and C.~L. {Giles}, ``Learning
  long-term dependencies in narx recurrent neural networks,'' \emph{IEEE
  Transactions on Neural Networks}, vol.~7, no.~6, pp. 1329--1338, Nov 1996.

\bibitem{zeevi}
A.~J. Zeevi, R.~Meir, and R.~J. Adler, ``Time series prediction using mixtures
  of experts,'' in \emph{NIPS}, 1996.

\bibitem{liehr}
S.~{Liehr}, K.~{Pawelzik}, J.~{Kohlmorgen}, S.~{Lemm}, and K.~R. {Muller},
  ``Hidden markov mixtures of experts for prediction of non-stationary
  dynamics,'' in \emph{Neural Networks for Signal Processing IX: Proceedings of
  the 1999 IEEE Signal Processing Society Workshop (Cat. No.98TH8468)}, 1999,
  pp. 195--204.

\bibitem{puma}
W.~J. {Puma-Villanueva}, C.~A.~M. {Lima}, E.~P. {dos Santos}, and F.~J. {Von
  Zuben}, ``Mixture of heterogeneous experts applied to time series: a
  comparative study,'' in \emph{Proceedings. 2005 IEEE International Joint
  Conference on Neural Networks, 2005.}, vol.~2, 2005, pp. 1160--1165 vol. 2.

\bibitem{ebrahimpour}
R.~Ebrahimpour, H.~Nikoo, S.~Masoudnia, M.~Yousefi, and M.~S. Ghaemi, ``Mixture
  of mlp-experts for trend forecasting of time series: A case study of the
  tehran stock exchange,'' \emph{International Journal of Forecasting},
  vol.~27, pp. 804--816, 07 2011.

\bibitem{diversity}
\emph{Diversity in Classifier Ensembles}.\hskip 1em plus 0.5em minus
  0.4em\relax John Wiley \& Sons, Ltd, 2014, ch.~8, pp. 247--289.

\bibitem{ephraim}
Y.~{Ephraim} and N.~{Merhav}, ``Hidden markov processes,'' \emph{IEEE
  Transactions on Information Theory}, vol.~48, no.~6, pp. 1518--1569, 2002.

\bibitem{nystrup}
P.~Nystrup, H.~Madsen, and E.~Lindström, ``Long memory of financial time
  series and hidden markov models with time-varying parameters,'' \emph{Journal
  of Forecasting}, vol.~36, 01 2016.

\bibitem{kozat1}
S.~S. {Kozat} and A.~C. {Singer}, ``Universal switching linear least squares
  prediction,'' \emph{IEEE Transactions on Signal Processing}, vol.~56, no.~1,
  pp. 189--204, 2008.

\bibitem{kozat2}
S.~S. Kozat and A.~C. Singer, ``Universal randomized switching,'' \emph{Trans.
  Sig. Proc.}, vol.~58, no.~3, p. 1922–1927, Mar. 2010.

\bibitem{changjin}
C.-J. Kim, ``Dynamic linear model with markov switching,'' \emph{Journal of
  Econometrics}, vol.~60, pp. 1--22, 02 1991.

\bibitem{hamilton}
J.~D. Hamilton, ``A new approach to the economic analysis of nonstationary time
  series and the business cycle,'' \emph{Econometrica}, vol.~57, no.~2, pp.
  357--384, 1989.

\bibitem{kns}
C.-J. Kim, C.~R. Nelson, and R.~Startz, ``Testing for mean reversion in
  heteroskedastic data based on gibbs-sampling-augmented randomization,''
  \emph{Journal of Empirical Finance}, vol.~5, no.~2, pp. 131 -- 154, 1998.

\bibitem{tvtp}
A.~J. Filardo, ``Business-cycle phases and their transitional dynamics,''
  \emph{Journal of Business \& Economic Statistics}, vol.~12, no.~3, pp.
  299--308, 1994.

\bibitem{markovbook}
S.~Frühwirth-Schnatter, ``Finite mixture and markov switching models,'' 01
  2006.

\bibitem{lstm}
S.~Hochreiter and J.~Schmidhuber, ``Long short-term memory,'' \emph{Neural
  computation}, vol.~9, pp. 1735--80, 12 1997.

\bibitem{gru}
K.~Cho, B.~van Merriënboer, C.~Gulcehre, F.~Bougares, H.~Schwenk, and
  Y.~Bengio, ``Learning phrase representations using rnn encoder-decoder for
  statistical machine translation,'' 06 2014.

\bibitem{Elman90}
J.~L. Elman, ``Finding structure in time,'' \emph{Cognitive Science}, vol.~14,
  no.~2, pp. 179--211, 1990.

\bibitem{Specht1991}
D.~F. {Specht}, ``A general regression neural network,'' \emph{IEEE
  Transactions on Neural Networks}, vol.~2, no.~6, pp. 568--576, Nov 1991.

\bibitem{Gou2007}
J.~Y. {Goulermas}, P.~{Liatsis}, X.~{Zeng}, and P.~{Cook}, ``Density-driven
  generalized regression neural networks (dd-grnn) for function
  approximation,'' \emph{IEEE Transactions on Neural Networks}, vol.~18, no.~6,
  pp. 1683--1696, Nov 2007.

\bibitem{Narendra90}
K.~S. {Narendra} and K.~{Parthasarathy}, ``Identification and control of
  dynamical systems using neural networks,'' \emph{IEEE Transactions on Neural
  Networks}, vol.~1, no.~1, pp. 4--27, March 1990.

\bibitem{tolga19}
T.~{Ergen} and S.~S. {Kozat}, ``Unsupervised anomaly detection with lstm neural
  networks,'' \emph{IEEE Transactions on Neural Networks and Learning Systems},
  pp. 1--15, 2019.

\bibitem{yang}
W.~Hu, Y.~Yang, Z.~You, Z.~Liu, and X.~Ren, ``Modeling combinatorial evolution
  in time series prediction,'' 05 2019.

\bibitem{kim_book}
C.-J. Kim and C.~Nelson, \emph{State-Space Models with Regime Switching:
  Classical and Gibbs-Sampling Approaches with Applications}, 01 2017.

\bibitem{kim}
M.~J. Kim, C.~R. Nelson, and R.~Startz, ``{Mean Reversion in Stock Prices? A
  Reappraisal of the Empirical Evidence},'' \emph{The Review of Economic
  Studies}, vol.~58, no.~3, pp. 515--528, 05 1991.

\bibitem{cao}
Y.~{Cao}, Y.~{Li}, S.~{Coleman}, A.~{Belatreche}, and T.~M. {McGinnity},
  ``Adaptive hidden markov model with anomaly states for price manipulation
  detection,'' \emph{IEEE Transactions on Neural Networks and Learning
  Systems}, vol.~26, no.~2, pp. 318--330, 2015.

\bibitem{Williams89}
\BIBentryALTinterwordspacing
R.~J. Williams and D.~Zipser, ``A learning algorithm for continually running
  fully recurrent neural networks,'' \emph{Neural Comput.}, vol.~1, no.~2, pp.
  270--280, Jun. 1989. [Online]. Available:
  \url{http://dx.doi.org/10.1162/neco.1989.1.2.270}
\BIBentrySTDinterwordspacing

\bibitem{hmm_book}
L.~R. {Rabiner}, ``A tutorial on hidden markov models and selected applications
  in speech recognition,'' \emph{Proceedings of the IEEE}, vol.~77, no.~2, pp.
  257--286, 1989.

\bibitem{laplace}
S.~Kotz, T.~Kozubowski, and K.~Podgorski, \emph{The Laplace Distribution and
  Generalizations}, 01 2001.

\bibitem{ZinkOGD}
M.~Zinkevich, ``Online convex programming and generalized infinitesimal
  gradient ascent,'' vol.~2, 04 2003.

\bibitem{Will95}
R.~J. Williams and D.~Zipser, ``Backpropagation,'' Y.~Chauvin and D.~E.
  Rumelhart, Eds.\hskip 1em plus 0.5em minus 0.4em\relax Hillsdale, NJ, USA: L.
  Erlbaum Associates Inc., 1995, ch. Gradient-based Learning Algorithms for
  Recurrent Networks and Their Computational Complexity, pp. 433--486.

\bibitem{data}
``Historical data feed,'' [Online]. Available:
  \url{https://www.dukascopy.com\\/swiss/english/marketwatch/historical/}
  [Accessed: 2020-01-24].

\end{thebibliography}

\end{document}